\documentclass{article}

\usepackage[preprint]{neurips_2026}

\usepackage[utf8]{inputenc} 
\usepackage[T1]{fontenc}    
\usepackage{hyperref}       
\usepackage{url}            
\usepackage{booktabs}       
\usepackage{amsfonts}       
\usepackage{nicefrac}       
\usepackage{microtype}      
\usepackage{xcolor}         

\usepackage{graphicx} 
\usepackage{amssymb}
\usepackage{multirow}   
\usepackage{tabularx}
\usepackage{float}
\usepackage{array} 

\newcolumntype{C}{>{\centering\arraybackslash}X}
\usepackage{amsthm}
\usepackage{amsmath, amsthm, amssymb}
\usepackage{bm}

\newtheorem{theorem}{Theorem}
\newtheorem{proposition}{Proposition}

\newtheorem{innercustomthm}{Theorem}
\newenvironment{customthm}[1]
  {\renewcommand\theinnercustomthm{#1}\innercustomthm}
  {\endinnercustomthm}

\title{SF-AMS: Strategic Forgetting for Structured Memory in LLM Agent}

\author{
  Ning Yang, Siqi Li, Miaoxin Shen, Yuan Zhou, Meng Zhang, Tong Li, Haijun Zhang
}

\begin{document}

\maketitle

\begin{abstract}
Managing long-context dependencies remains a primary bottleneck in LLM agents, as redundant and irrelevant information can degrade multi-step reasoning. Strategic Forgetting for Agent Memory Systems (SF-AMS) is proposed as a framework for maintaining compact high-utility memory by modeling the long-term importance of memory units. SF-AMS replaces static retrieval and heuristic decay with a utility-driven survival mechanism that updates memory importance from usage redundancy and temporal signals, inducing a hierarchical memory structure that prioritizes stable entity-consistent information while filtering noise. On top of this, Composite Importance Scoring integrates semantic and entity level signals to improve retrieval robustness. Experiments on LoCoMo and LongMemEval-s show consistent gains over strong state of the art baselines including LightMem MemO and A-Mem. The largest improvement appears in multi-hop reasoning under Qwen2.5-7B where SF-AMS achieves plus 9.65 F1 over the strongest baseline followed by temporal reasoning under GPT-4o-mini plus 6.91 F1 and open-domain tasks plus 6.53 F1 demonstrating strong cross backbone generalization. These results show that modeling memory importance as a dynamic utility signal is critical for reliable long-context reasoning.
\end{abstract}

\section{Introduction}
Large Language Model (LLM) agents have demonstrated remarkable reasoning and decision-making capabilities in complex, real-world environments \citep{xi2025rise, zhang2025survey, Du2025RethinkingMI, Mei2024AIOSLA, wang2023scm}. Sustaining performance over long-horizon interactions remains challenging, as memory systems must retain essential information while limiting the accumulation of noisy context. Recent empirical studies show that increasing context length alone does not guarantee effective utilization of relevant information, and models often do not pay attention to important signals in long sequences \citep{liu2023lostmiddle}. Despite significant progress, most existing memory systems remain dependent on predefined storage structures and fixed retrieval strategies. Even graph-based approaches \citep{edge2024graphrag, rasmussen2025zep} and hierarchy-based methods \citep{packer2023memgpt, kang2025memoryos} are often constrained by rigid schemas, limiting their ability to dynamically reorganize knowledge and form new conceptual connections over time. As a result, these systems often suffer from a declining signal-to-noise ratio, which hinders their ability to generalize to new tasks and maintain reasoning consistency over extended periods.

Previous research has addressed these challenges through three primary paradigms. 
(1) \textit{Knowledge organization methods} \citep{xu2025a, chhikara2025mem0, Modarressi2023RETLLMTA, rasmussen2025zep,  salama2025meminsight}, such as A-Mem and Mem0, structure historical data into atomic units or relational graphs to improve interpretability and support structured reasoning. 
(2) \textit{Retrieval augmented and memory access approaches} \citep{zhong2024memorybank, wang2025scm, jiang2023flare, lin2024radit, tang2026activerag}, exemplified by MemoryBank, model memory access through retrieval, time decay, or iterative reasoning mechanisms to improve information utilization. 
(3) \textit{System level and dynamic memory management frameworks} \citep{packer2023memgpt, kang2025memoryos, liu2024agentlite, wang2025openhands, fang2025lightmem,  yu2026agentic}, including MemGPT and MemoryOS, introduce explicit control over memory storage and updates through hierarchical structures and scheduling strategies. Despite their effectiveness, these approaches typically address either memory organization or retrieval in isolation, lacking a unified mechanism that continuously balances transient interaction noise against long-term utility.

To address this gap, Strategic Forgetting for Agent Memory Systems (SF-AMS) is proposed as an agentic memory framework that dynamically maintains a compact and high-utility memory state. Unlike passive storage systems, SF-AMS actively tracks the utility of each memory unit and updates it over time based on usage, redundancy, and temporal signals. A four-layer hierarchical structure prioritizes critical information, while an entity anchoring mechanism preserves factual consistency. Redundant or low-value memories are gradually removed through a controlled decay process, preventing noise accumulation. As a result, the system maintains a compact yet informative memory state, enabling more stable reasoning under limited context budgets. The contributions are summarized as follows:
\begin{itemize}
    \item SF-AMS is proposed as a dynamic memory framework that explicitly controls memory growth through strategic forgetting. By continuously filtering redundant and low-utility information, the system maintains a compact memory state and mitigates noise accumulation during long-horizon interactions.

    \item A utility-driven memory update mechanism is introduced, which assigns each memory unit a dynamic score based on usage frequency, redundancy, and temporal signals. This mechanism reinforces frequently accessed and informative memories while gradually removing less useful ones, enabling adaptive and self-organizing memory evolution.

    \item Experiments on LoCoMo and LongMemEval-s demonstrate consistent improvements over strong baselines, achieving up to +9.65 F1 in multi-hop reasoning and +6.91 F1 in temporal reasoning, indicating robust performance across reasoning types and backbone models.
\end{itemize}

\section{Related Work}

\subsection{Knowledge Organization Methods}

Knowledge organization methods aim to improve memory quality by transforming raw interactions into structured and interpretable representations \citep{xu2025a, chhikara2025mem0, Modarressi2023RETLLMTA}. Early approaches organize memory as atomic notes or relational structures to support incremental accumulation of knowledge and structured reasoning \citep{xu2025a, chhikara2025mem0}. Subsequent work explores more expressive representations, including relational triplets, attribute-enhanced memory, and grounded memory systems that integrate perception and semantic context \citep{Modarressi2023RETLLMTA, salama2025meminsight, ocker2025grounded}. More recent studies further extend memory to graph-based and temporal structures, enabling global reasoning and cross-session consistency \citep{rasmussen2025zep, edge2024graphrag}. In addition, alternative paradigms propose the storage of intermediate reasoning traces instead of raw text to improve reuse and consistency \citep{liu2023tim}. Recent work also introduces more compressed or segmented memory representations to mitigate redundancy and improve retrieval efficiency \citep{Pan2025OnMC}. Despite these advances, most methods rely on continuous accumulation, which often leads to redundancy and reduced efficiency over long interaction horizons. To address this limitation, our method introduces a dynamic memory regulation mechanism that explicitly evaluates memory utility and redundancy, enabling adaptive retention of informative content while suppressing unnecessary accumulation.

\subsection{Retrieval Augmented and Memory Access Approaches}

Retrieval augmented approaches enhance LLM reasoning by dynamically incorporating external knowledge and selectively accessing memory \citep{lewis2020rag, jiang2023flare, trivedi2023ircot, lin2024radit}. Standard retrieval pipelines rely on semantic matching between queries and indexed memory, while more advanced methods introduce iterative retrieval-generation loops and self-reflective mechanisms to improve reasoning quality \citep{trivedi2023ircot}. Other approaches focus on optimizing retrieval decisions through uncertainty estimation, joint retriever-generator training, or document-level evaluation \citep{jiang2023flare, lin2024radit, tang2026activerag, yu2024chainofnote}. 
Iterative reasoning frameworks further improve performance by interleaving retrieval and generation steps to support multi-hop reasoning. In addition, recent studies explore rationally aware efficiency mechanisms that dynamically control the depth or necessity of reasoning steps, including skipping redundant computation, compressing intermediate reasoning traces, and adapting reasoning complexity to query difficulty \citep{zhang2025ascot, zhang2026chain, zhang2026not}. In parallel, memory-oriented retrieval strategies incorporate time-decay or usage signals to simulate human-like recall \citep{zhong2024memorybank}. 
However, these methods primarily focus on retrieval accuracy and access strategies, while largely overlooking redundancy control and long-term memory evolution. To overcome this limitation, our approach integrates retrieval with adaptive memory filtering and evolution, enabling the model to jointly optimize retrieval relevance and long-term memory efficiency under a unified control mechanism.

\subsection{System Level and Dynamic Memory Management Frameworks}

The system level and dynamic memory management frameworks address long-context limitations through structured architectures and adaptive control mechanisms \citep{packer2023memgpt, kang2025memoryos, Mei2024AIOSLA}. Early designs introduce virtual memory abstractions and hierarchical storage to manage context constraints \citep{packer2023memgpt, kang2025memoryos}, while later systems incorporate OS-inspired scheduling and modular agent infrastructures \citep{Mei2024AIOSLA, liu2024agentlite, wang2025openhands}. Other approaches explore controller-based memory management and cognitive-inspired consolidation strategies to balance efficiency and retention \citep{wang2023scm, fang2025lightmem}. More recent work investigates learning-based memory updates, compression techniques, and reinforcement learning for adaptive memory control \citep{Pan2025OnMC, behrouz2025titans, yu2026agentic}, further moving towards learnable policies for memory writing and forgetting. Personalized memory systems further incorporate user-specific signals into long-term interactions to improve consistency across sessions \citep{li2025hello}. Nevertheless, these frameworks often rely on predefined rules or single optimization signals, limiting their ability to jointly model importance, redundancy, and long-term utility. To address this issue, our framework introduces a unified strategic forgetting mechanism that dynamically balances memory retention and removal based on multi-dimensional utility signals, enabling more robust long-term memory management.

\section{Methodology}

\subsection{Framework Philosophy and System Architecture}

\begin{figure*}[t]
    \centering
    \includegraphics[width=\textwidth]{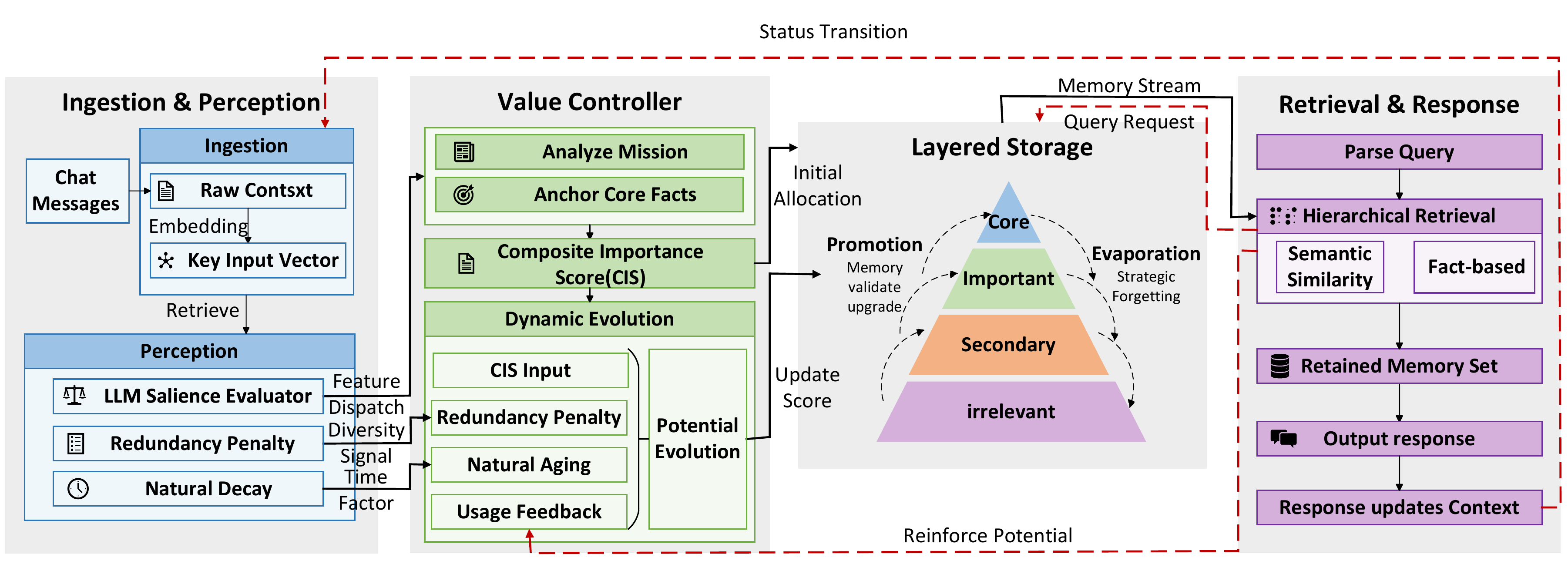}
    \caption{
    Overview of the proposed SF-AMS framework. The system consists of four stages: (1) ingestion and perception, where raw conversational inputs are transformed into structured signals; (2) value control, where Composite Importance Scoring (CIS) evaluates memory salience; (3) layered storage, where memories are organized into a hierarchical structure and dynamically updated via survival potential evolution; and (4) retrieval and response, where queries are resolved through hybrid retrieval mechanisms combining semantic similarity and fact-based reasoning. The entire pipeline forms a closed-loop memory stream, enabling continuous refinement of the memory state.
    }
    \label{fig:framework}
\end{figure*}

The proposed Strategic Forgetting Agent Memory System (SF-AMS) is built on the insight that long-term adaptability depends on selectively retaining information with high structural utility rather than increasing storage capacity. Existing LLM-based memory systems often exhibit a linear accumulation effect, where uncontrolled growth degrades the signal-to-noise ratio over long interaction horizons. To address this, SF-AMS models memory as a dynamic system in which each unit evolves according to its survival potential. The framework comprises three components: a perceptual encoding module, a survival-driven memory store, and a value-guided controller, which jointly govern retention and eviction under capacity constraints.

As illustrated in Figure~\ref{fig:framework}, SF-AMS is implemented as a four-stage pipeline: ingestion and perception, value-guided control, hierarchical storage, and retrieval-response integration. Inputs are encoded into key vectors enriched with salience, redundancy, and temporal signals, which are evaluated using the Composite Importance Scoring (CIS) mechanism to estimate structural importance. The memory module maintains a hierarchical organization (Core, Important, Secondary, Irrelevant), where each unit evolves under memory evolution dynamics driven by redundancy penalties, decay, and usage feedback. During inference, queries are resolved via hybrid retrieval that combines semantic similarity with fact-aware matching, and the retrieved memory supports response generation, which is re-integrated to continuously refine the memory structure.

\subsection{Problem Formulation and State-Space Modeling}

Memory management is formulated as a discrete-time state evolution process. At each time step $t$, the system is characterized by a state $s_t$ consisting of four components: short-term contextual input $C_t$, stored memory set $M_t$, utility signals $\Phi_t$, and task context $T$. The task context $T$ is assumed to remain fixed within an interaction episode, while $C_t$, $M_t$, and $\Phi_t$ evolve dynamically over time as new information is observed and incorporated into the system.

The memory set is defined as $M_t = \{m_1, \dots, m_{n_t}\}$, where each memory unit $m_i$ represents an interaction, a factual statement, or an extracted information element. Memory size is constrained by a fixed capacity $|M_t| \leq N_{\max}$, and the number of newly inserted memory units at each step is bounded by a small constant determined by the interaction granularity. To support structured memory evaluation under these constraints, each memory unit $m_i$ is associated with a scalar survival potential $\Phi(m_i)$, and the collection of these values is denoted as $\Phi_t = \{\Phi(m_1), \dots, \Phi(m_{n_t})\}$.

The survival potential $\Phi(m_i)$ reflects the relative utility of each memory unit and is updated dynamically based on factors such as usage frequency, redundancy with existing memories, and temporal decay. Intuitively, frequently accessed or highly informative memory units tend to accumulate higher survival potential, while redundant or rarely used ones gradually lose importance. Based on this formulation, SF-AMS adopts a utility-driven memory maintenance strategy in which $\Phi(m_i)$ serves as a unified criterion for both retrieval prioritization and memory pruning under capacity constraints. This enables the system to maintain a compact yet informative memory state over time without relying on static heuristics or fixed retrieval policies.

\subsection{Composite Importance Scoring (CIS) and Entity Anchoring}

SF-AMS introduces a Composite Importance Scoring (CIS) mechanism that maps heterogeneous memory signals into a unified importance score, which jointly governs retrieval and memory evolution. The importance score for each memory unit $m_i$ is defined as
\begin{equation}
I(m_i) = \text{Norm}\Bigl(\alpha S_{\text{LLM}}(m_i) + (1-\alpha) \sum_{k \in K} \mathbf{1}_k(m_i) w_k \Bigr),
\label{eq:2}
\end{equation}
where $S_{\text{LLM}}(m_i)$ is a task-conditioned salience score, $\mathbf{1}_k(m_i)$ indicates the presence of entity type $k$, and $w_k$ encodes structural priors. The parameter $\alpha \in [0,1]$ balances semantic and entity-aware signals.

The normalization operator maps raw importance scores $x_i$ over the current memory set $M_t$ into a bounded range:
\begin{equation}
\text{Norm}(x_i) =
\frac{x_i - \min_{m_j \in M_t} x_j}{\max_{m_j \in M_t} x_j - \min_{m_j \in M_t} x_j + \epsilon},
\end{equation}
where $x_i$ denotes the unnormalized importance score of the memory unit $m_i$, and the minimum and maximum are computed over all memory units in the current set $M_t$. The small constant $\epsilon > 0$ ensures numerical stability. This operation produces normalized scores in $[0,1]$, enabling stable comparison under dynamically evolving memory distributions. Based on $I(m_i)$, a ranking function $r(m_i)=\text{rank}(I(m_i))$ induces a total ordering over $M_t$, which is partitioned into hierarchical layers $\mathcal{L}=\{L_1,\dots,L_4\}$. This hierarchy directly determines Top-$K$ retrieval in Eq.~\ref{eq:5} and is also used as a control signal for survival updates, coupling retrieval and long-term memory persistence under a unified mechanism.

This ranking structure also enables a formal guarantee on retrieval stability. When sufficient separation exists between core and distractor memories, CIS induces robust ordering under perturbations. The following theorem formalizes this condition:
\begin{theorem}[Hierarchical Retrieval Priority]
\label{theorem1}
Under the structural gap condition $\gamma > 1 + \delta$, core factual invariants in $L_1$ are strictly ranked above distractors in $L_4$ under Eq.~\ref{eq:5}, ensuring stable Top-$K$ retrieval.
\end{theorem}
The detailed proof of this theorem is provided in Appendix~\ref{appendix:theorem1}. The condition $\gamma > 1 + \delta$ guarantees that perturbations cannot reverse the CIS-induced ordering, thereby preserving hierarchical separation throughout memory updates.

\subsection{Survival Potential Dynamics and Strategic Forgetting}

The survival dynamics in Eq.~\ref{eq:3} are controlled by the CIS output $I(m_i)$, which determines the reinforcement strength in the additive update of the survival potential $\Phi_i^t$. Accordingly, the evolution of each memory unit is formulated as a discrete-time state transition:
\begin{equation}
\Phi_i^{t+1} = \max\{0, \Phi_i^t + I(m_i)\cdot \Gamma_{\text{usage}}^t \cdot \Psi_{\text{div}}(m_i, M_t) - \lambda\},
\label{eq:3}
\end{equation}
where $\Phi_i^t \ge 0$ denotes the survival potential of memory unit $m_i$, $I(m_i)$ is the CIS-derived importance score (Eq.~\ref{eq:2}), and $\lambda > 0$ is a global decay coefficient controlling temporal forgetting. The binary usage indicator $\Gamma_{\text{usage}}^t$ and diversity modulation $\Psi_{\text{div}}(m_i, M_t)$ jointly act as gating factors for reinforcement.

The diversity modulation is defined as:
\begin{equation}
\Psi_{\text{div}} = 1 + \alpha_{\text{div}} \tanh\big(\beta(\tau - \text{Sim}(m_i, M_t))\big),
\label{eq:4}
\end{equation}
where $\text{Sim}(m_i, M_t) \in [0,1]$ measures the maximum semantic similarity between $m_i$ and elements in $M_t$, $\alpha_{\text{div}} \in (0,1)$ controls the modulation strength, $\tau \in [0,1]$ defines the similarity threshold, and $\beta > 0$ controls the transition sharpness. This formulation adaptively modulates reinforcement based on semantic redundancy: highly similar memories are down-weighted, while less similar or novel memories are amplified through a smooth gating effect.

To ensure long-term stability, it is necessary to characterize the global behavior induced by Eq.~\ref{eq:3}, since uncontrolled reinforcement may otherwise lead to divergence. The multiplicative structure in Eq.~\ref{eq:3} acts as a gated update mechanism where reinforcement is activated only when importance, usage, and diversity jointly contribute, while the decay term $\lambda$ enforces contraction. 

\begin{proposition}[Boundedness and Stability]
\label{proposition1}
Under the memory dynamics defined in Eq.~\ref{eq:3}, and assuming bounded reinforcement and capacity-constrained updates, the aggregate survival potential $V(X_t)$ remains uniformly bounded over time.
\end{proposition}

The detailed proof of this proposition is provided in Appendix~\ref{appendix:A2}. This result implies that the memory system evolves within a stable regime where reinforcement and decay remain balanced over time, preventing both uncontrolled growth and collapse of memory states. In particular, it ensures that the CIS-driven reinforcement mechanism does not induce divergence even under repeated reinforcement of high importance memory units.

\subsection{Hierarchical Multi-Source Retrieval Pipeline}

The survival potential $\Phi_i^t$ is incorporated into the retrieval stage as a ranking prior that biases memory selection toward long-term high-utility units. Memory retrieval follows a hierarchical multi-source pipeline that combines coarse-to-fine matching to improve both efficiency and precision. Given a query $q$, structural parsing first extracts keywords, entities, and temporal expressions, along with optional query variants $\mathcal{Q}_{var}$ to reduce lexical mismatch. Retrieval is then performed in a staged manner across multiple memory levels. 

For a candidate session $S_j$, defined as a temporally contiguous sequence of memory units, relevance is computed as:
\begin{equation}
S_{\text{session}}(q, S_j) = \text{sim}_{\text{sem}}(q, S_j) + \alpha_{\text{kw}} \cdot \text{sim}_{\text{kw}}(q, S_j),
\label{eq:5}
\end{equation}
where $\text{sim}_{\text{sem}}$ measures embedding-based semantic similarity, $\text{sim}_{\text{kw}}$ evaluates keyword overlap, and $\alpha_{\text{kw}}$ balances the two signals. 
Sessions exceeding a threshold are further expanded into fine-grained memory-level ranking, where lexical, semantic, temporal, and entity-aware signals are jointly integrated to compute relevance scores. This produces the final retrieved set $M_t^*$, defined as the ordered and deduplicated collection of selected memory units at time $t$. The retrieved set $M_t^*$ is then fed back into the memory system by marking all selected memory units as accessed, which directly updates the usage indicator $\Gamma_{\text{usage}}^t$ in Eq.~\ref{eq:3}. This establishes a closed feedback loop in which retrieval decisions influence future survival dynamics, as frequently accessed memories receive higher reinforcement through repeated updates of the survival potential.

\section{Experiments}

\subsection{Experimental Setup}\label{sec:experimental_setup}

\textbf{Datasets.} 
The proposed framework is evaluated on two representative benchmarks: LoCoMo \citep{maharana2024locomo} and LongMemEval-s \citep{wu2025longmemeval}. LoCoMo contains long-context conversational data with an average of 300 turns and 9K tokens per dialogue, enabling evaluation across single-hop, multi-hop, temporal, and open-domain reasoning. LongMemEval-s provides a large-scale benchmark with 14K sessions and 146K turns, focusing on information extraction and cross-session knowledge updates.

\textbf{Evaluation Metrics.} 
Performance is evaluated using benchmark-specific metrics. For LoCoMo, we report F1 and BLEU-1 \citep{papineni-etal-2002-bleu} to measure semantic and lexical matching quality. For LongMemEval-s, Accuracy (ACC) is used to assess structured knowledge extraction. System efficiency is additionally measured in terms of token usage, API overhead, and runtime.

\textbf{Baselines.} 
SF-AMS is compared against representative memory-augmented systems, including TiM \citep{liu2023tim}, MemoryBank \citep{zhong2024memorybank}, MemGPT \citep{packer2023memgpt}, A-Mem \citep{xu2025a}, MemoryOS \citep{kang2025memoryos}, and LightMem \citep{fang2025lightmem}. These methods cover a range of strategies such as trace-based retrieval, temporal decay modeling, OS-inspired memory management, adaptive memory structuring, and memory compression.

\subsection{Implementation Details}\label{sec:implementation_details}

The memory system maintains a fixed capacity of $N_{\max}=300$, enforcing a bounded memory budget. For retrieval, the 7 most recent interaction-derived memory notes are prioritized, while the remaining entries are stored in the global memory set $\mathcal{M}$. Each memory unit is encoded using the all-MiniLM-L6-v2 encoder, with cosine similarity adopted for semantic matching. During inference, the top-$k$ ($k=10$) most relevant memory units are selected using the hybrid scoring function in Eq.~\ref{eq:5}, which combines semantic similarity and keyword matching via a fixed coefficient $\alpha_{\text{kw}}$. Memory importance is computed using the Composite Importance Scoring (CIS) in Eq.~\ref{eq:2}, while survival potential and redundancy modulation follow Eq.~\ref{eq:3} and Eq.~\ref{eq:4}, respectively. All scoring components are normalized or inherently bounded, helping maintain stable reinforcement dynamics throughout the update process.

\subsection{Main Results}

\begin{table}[t]
\caption{Performance comparison on the LoCoMo benchmark using F1 and BLEU-1 metrics. Our method achieves strong and consistent improvements across most reasoning types and both backbone models, particularly in single-hop, open-domain, and temporal tasks, demonstrating robust cross-backbone generalization. The best results are shown in bold, and the second-best results are underlined.}
\centering
\small
\setlength{\tabcolsep}{4pt}

\begin{tabular*}{\textwidth}{@{\extracolsep{\fill}}llcccccccc}
\toprule
\multirow{2}{*}{Model} & \multirow{2}{*}{Method} 
& \multicolumn{2}{c}{Single-Hop} 
& \multicolumn{2}{c}{Multi-Hop} 
& \multicolumn{2}{c}{Open Domain} 
& \multicolumn{2}{c}{Temporal} \\
& 
& F1 & BLEU-1 
& F1 & BLEU-1 
& F1 & BLEU-1 
& F1 & BLEU-1 \\
\midrule

\multirow{7}{*}{GPT-4o-mini}
& Long-context & 20.88 & 13.94 & 20.97 & 13.74 & 4.88 & 2.41 & 14.46 & 9.39 \\
& TiM        & 16.25 & 13.12 & 18.43 & 17.35 & 23.74 & 22.05 & 8.35 & 7.32 \\
& MemoryBank & 5.00  & 4.77  & 9.68  & 6.99  & 6.61  & 5.16  & 5.56 & 5.94 \\
& MemGPT     & 26.65 & 17.72 & 25.52 & 19.44 & 41.04 & 34.34 & 9.15 & 7.44 \\
& A-Mem      & 27.02 & 20.09 & \textbf{45.85} & \textbf{36.67} & 44.65 & 37.06 & 12.14 & 12.00 \\
& MemO       & \underline{35.27} & \underline{25.22} & \underline{41.15} & 30.76 & \underline{48.62} & \underline{42.99} & 20.02 & 16.52 \\
& LightMem & 26.16 & 22.26 & 30.76 & 21.90 & 23.95 & 13.16 & \underline{38.93} & \underline{33.64} \\
& \textbf{Ours} 
& \textbf{40.21} & \textbf{34.40} 
& 40.57 & \underline{31.87} 
& \textbf{49.10} & \textbf{47.13} 
& \textbf{45.84} & \textbf{38.98} \\
\midrule

\multirow{7}{*}{Qwen2.5-7B}
& Long-context & 13.49 & 8.73 & 14.68 & 9.59 & 5.71 & 3.49 & 6.47 & 4.11 \\
& TiM        & 4.37 & 5.01 & 2.54 & 3.21 & 6.35 & 7.32 & 6.20 & 5.37 \\
& MemoryBank & 3.60 & 3.39 & 1.72 & 1.97 & 4.11 & 3.32 & 6.63 & 6.58 \\
& MemGPT     & 5.07 & 4.31 & 2.96 & 2.95 & 7.26 & 5.52 & 7.04 & 7.10 \\
& A-Mem      & 12.57 & 9.01 & 27.59 & \underline{25.07} & 17.26 & 13.12 & 7.12 & 7.28 \\
& MemO       & 23.26 & 15.39 & 21.44 & 14.95 & \underline{26.23} & \underline{22.39} & 10.18 & 8.81 \\
& LightMem & \underline{25.67} & \textbf{21.85} & \underline{27.94} & 21.26 & 10.12 & 5.74 & \underline{24.97} & \underline{20.72} \\
& \textbf{Ours} 
& \textbf{26.70} & \underline{21.66} 
& \textbf{37.59} & \textbf{30.69}
& \textbf{32.76} & \textbf{29.48} 
& \textbf{29.24} & \textbf{25.73} \\
\bottomrule
\end{tabular*}

\label{tab:locomo}
\end{table}

The results in Table~\ref{tab:locomo} show that our method achieves the most pronounced improvements in structurally challenging scenarios, rather than uniform gains across all settings. These results indicate that the proposed mechanism is particularly effective when reasoning requires resolving long-range dependencies and filtering noisy context.

Under the GPT-4o-mini backbone, the largest gain is observed in Temporal reasoning, where our method achieves 45.84 F1, surpassing the strongest baseline (LightMem) by +6.91 F1. It also improves Single-Hop performance to 40.21 F1, exceeding the best baseline by +4.94 F1. Under the Qwen2.5-7B backbone, the most significant improvement appears in Multi-Hop reasoning, reaching 37.59 F1 and outperforming the strongest baseline by +9.65 F1. Additionally, Open-Domain performance improves to 32.76 F1, with a gain of +6.53 F1. These results show consistent generalization across different backbones, particularly in scenarios requiring accurate memory selection.

These improvements can be attributed to the structure-aware ranking induced by the Composite Importance Scoring (CIS) mechanism. By jointly modeling semantic relevance and entity-level signals, the method forms a more separable memory hierarchy that reduces retrieval interference. This behavior aligns with Theorem~\ref{theorem1}, which guarantees that structurally important memories are consistently ranked above distractors, thereby improving retrieval reliability in complex reasoning scenarios.

\subsection{Ablation Study}

\begin{table}[t]
\centering
\small
\setlength{\tabcolsep}{6pt}
\renewcommand{\arraystretch}{1.15}

\caption{Ablation study on the LoCoMo dataset using the Qwen2.5-7B backbone. Results show that all proposed components contribute significantly to overall performance, with Semantic Salience and Entity Anchoring being the most critical modules. The best results are shown in bold, and the second-best results are underlined.}
\label{tab:ablation_table}

\begin{tabular*}{\textwidth}{@{\extracolsep{\fill}}lcccc}
\toprule
\textbf{Variant} & \textbf{Single-Hop} & \textbf{Multi-Hop} & \textbf{Open Domain} & \textbf{Temporal} \\
\midrule

\textbf{SF-AMS (Full)} & \textbf{26.70} & \textbf{37.59} & \textbf{32.76} & \textbf{29.24} \\
w/o Semantic Salience   & \underline{21.70} & 14.60 & 10.60 & \underline{18.80} \\
w/o Entity Anchoring    & 16.90 & 13.50 & 11.80 & 16.20 \\
w/o Strategic Forgetting & 16.11 & \underline{18.78} & \underline{14.06} & 15.16 \\
w/o Memory System       & 9.60  & 14.25 & 13.69 & 5.27 \\
\bottomrule
\end{tabular*}

\end{table}

\begin{figure*}[t]
\centering
\includegraphics[width=\textwidth]{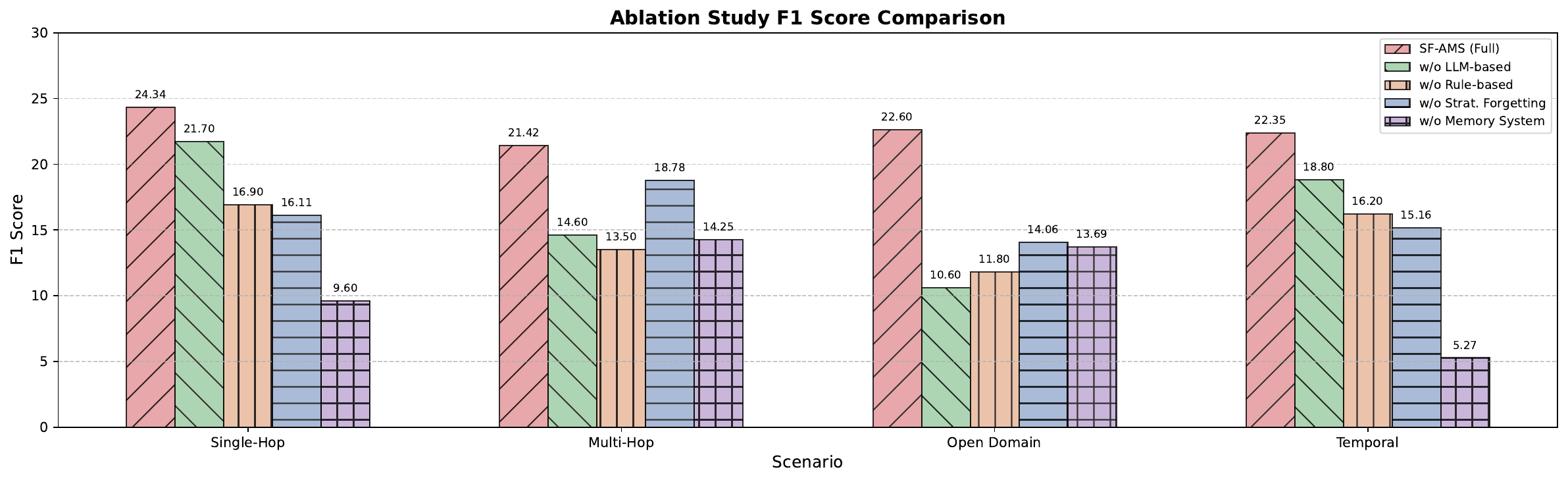}
\caption{Ablation results across four reasoning categories on the LoCoMo dataset. The figure shows the F1 score degradation of each ablated variant relative to the full SF-AMS model, demonstrating the contribution of each module across different reasoning types.}
\label{fig:ablation_barplot}
\end{figure*}

To evaluate the contribution of each component, an ablation study is conducted by systematically removing key modules from SF-AMS, including semantic salience scoring, entity anchoring, strategic forgetting, and the full memory system. Results are reported in Table~\ref{tab:ablation_table} and Figure~\ref{fig:ablation_barplot}. Removing the entire memory system leads to the most severe degradation, particularly in temporal reasoning, where performance drops by 23.97 F1, confirming that a structured memory state is essential for modeling long-range dependencies.

At the component level, different modules exhibit complementary effects across reasoning types. Removing entity anchoring reduces multi-hop performance by 24.09 F1 and temporal performance by 13.04 F1, indicating the importance of explicit structural signals for cross-step dependency tracking. In contrast, removing semantic salience causes the largest drop in open-domain reasoning, with a decrease of 22.16 F1, highlighting the role of LLM-based scoring in handling unstructured queries. Furthermore, disabling strategic forgetting leads to consistent degradation across all categories, with reductions of 10.59 F1 in single-hop, 18.81 F1 in multi-hop, 18.70 F1 in open-domain, and 14.08 F1 in temporal reasoning. Overall, these results demonstrate that the proposed components contribute complementary functionalities, and their integration is necessary for stable and strong performance.

\subsection{Efficiency Analysis}

\begin{table*}[t]
\caption{Unified performance and efficiency comparison on the LongMemEval-s benchmark. We report token-level memory cost together with system-level efficiency and accuracy. The best results are shown in bold, and the second-best results are underlined.}
\centering
\small
\setlength{\tabcolsep}{3pt}
\renewcommand{\arraystretch}{1.1}

\begin{tabular*}{\textwidth}{@{\extracolsep{\fill}}lcccccccc}
\toprule
\textbf{Method} 
& \textbf{ACC (\%)} 
& \multicolumn{2}{c}{\textbf{Summary Tokens (k)}} 
& \multicolumn{2}{c}{\textbf{Update Tokens (k)}} 
& \textbf{Total (k)} 
& \textbf{LLM Calls} 
& \textbf{Runtime (s)} \\
\cmidrule(lr){3-4} \cmidrule(lr){5-6}
& 
& In & Out & In & Out & & & \\
\midrule

Full Text   & 54.80 & -- & -- & -- & -- & 105.07 & --      & -- \\
NaiveRAG    & 60.80 & -- & -- & -- & -- & --     & --      & 659.09 \\
LangMem     & 50.80 & -- & -- & 1311.96 & 118.06 & \underline{1430.02} & \underline{495.12}  & \underline{3237.16} \\
A-MEM       & \underline{65.20} & 219.21 & 66.98 & 1260.54 & 318.20 & 1864.93 & 989.30  & 5367.51 \\
MemoryOS    & 49.60 & 2102.54 & 510.88 & 305.12 & 27.43 & 2944.97 & 2922.28 & 8721.78 \\
Mem0        & 39.51 & 424.20 & 15.34 & 411.50 & 111.35 & \textbf{1001.90} & 722.76  & \textbf{2239.94} \\
\midrule

\textbf{Ours} 
& \textbf{68.52} 
& 1600.30 & 154.97 & -- & -- & 1755.27 
& \textbf{263.99} & 4323.39 \\
\bottomrule
\end{tabular*}

\label{tab:longmem_unified}
\end{table*}

The efficiency results on the LongMemEval-s benchmark are summarized in Table~\ref{tab:longmem_unified}. SF-AMS achieves the highest accuracy of 68.52\%, improving over the strongest baseline A-MEM by +3.32 ACC. At the same time, it significantly reduces LLM calls, achieving relative reductions of 46.7\% compared to LangMem and 73.3\% compared to A-MEM. These results indicate that SF-AMS improves reasoning performance while substantially lowering interaction overhead.

This improvement comes with moderate increases in system cost. SF-AMS incurs higher runtime than lightweight baselines such as Mem0 (2239.94 s) and LangMem (3237.16 s), reaching 4323.39 s, but remains substantially more efficient than heavier systems such as A-MEM (5367.51 s). In terms of token consumption, SF-AMS requires 1755.27k tokens, which is higher than Mem0 and LangMem but still comparable to existing memory-based methods. These results suggest that SF-AMS achieves improved reasoning performance without incurring excessive overhead. By prioritizing high-utility memory units and reducing redundant LLM interactions, the framework strikes a practical balance between accuracy and system efficiency.

\subsection{Memory Distribution Analysis}

\begin{table}[t]
\centering
\small

\caption{Distribution of memory instances across different importance levels and their corresponding forgetting ratios.}
\label{tab:memory_distribution}

\begin{tabular*}{\textwidth}{@{\extracolsep{\fill}}lcccc}
\toprule
\textbf{Level} & \textbf{Samples} & \textbf{Ratio} & \textbf{Avg. Importance} & \textbf{Forget Ratio} \\
\midrule
Core        & 305  & 10.62\% & 0.930 & 0.00\% \\
Important   & 409  & 14.25\% & 0.745 & 14.85\% \\
Secondary   & 1413 & 49.22\% & 0.546 & 39.83\% \\
Irrelevant  & 744  & 25.91\% & 0.356 & 64.74\% \\
\bottomrule
\end{tabular*}
\end{table}

To examine how the strategic forgetting mechanism shapes memory organization, the distribution across hierarchical importance levels is analyzed. As shown in Table~\ref{tab:memory_distribution}, the resulting pattern is highly skewed: only 10.62\% of memory units belong to the Core layer, yet they retain the highest average importance (0.930). This indicates that the Composite Importance Scoring (CIS) effectively identifies a compact set of stable reasoning anchors rather than distributing importance uniformly across the memory space. All statistics are aggregated over the full interaction trajectory, while the memory capacity is constrained by $N_{\max}=300$ in the main experiments.

\begin{figure}[H]
\centering
\includegraphics[width=\textwidth]{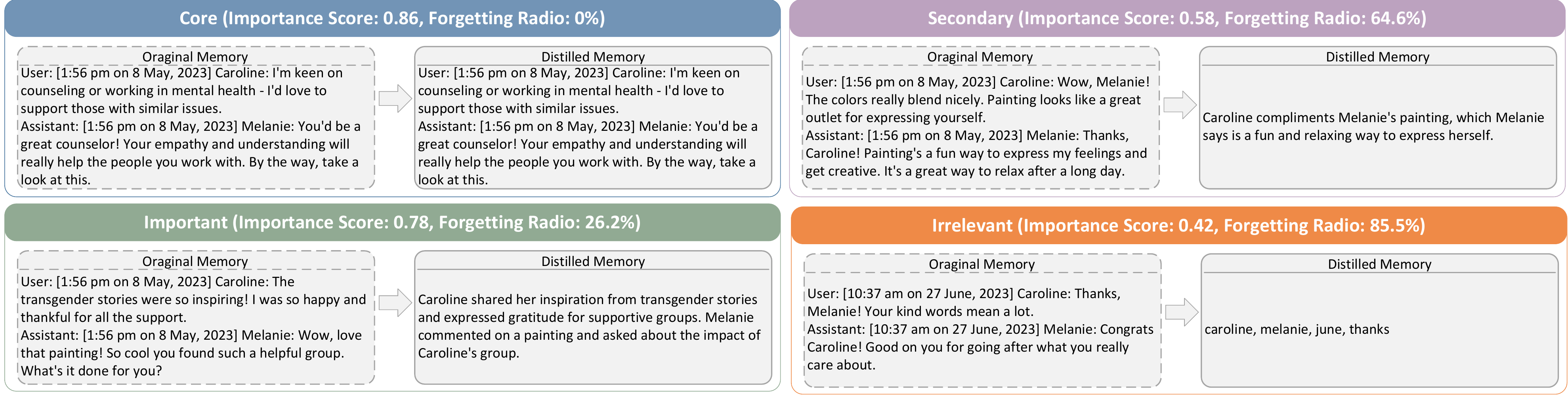} 
\caption{Visualizing the hierarchical memory distillation process in SF-AMS. We showcase one representative QA trace for each of the four levels: Core, Important, Secondary, and Irrelevant.}
\label{fig:forgetting_trace}
\end{figure}

A clear tiered forgetting pattern emerges from the dynamics. Core memories are rarely discarded, whereas Irrelevant memories are aggressively removed, with a forgetting ratio of 64.74\%. The Secondary layer (49.22\% of memory) undergoes moderate decay (39.83\%), functioning as a transitional buffer between stable knowledge and transient context. As illustrated in Figure~\ref{fig:forgetting_trace}, factual invariants are consistently preserved in higher layers, while redundant or low-signal dialogue turns are gradually discarded. This hierarchical organization enables SF-AMS to maintain a compact yet informative memory state with an improved signal-to-noise ratio.

\section{Conclusions}

This paper presents SF-AMS, a structured memory framework for long-term memory management in LLM agents. The proposed system combines composite importance scoring with survival-driven dynamics to enable selective retention and strategic forgetting, allowing memory to evolve in a controlled and adaptive manner. Experiments on LoCoMo and LongMemEval-s show that SF-AMS achieves consistent improvements in long-horizon reasoning while maintaining practical system efficiency. Compared to existing memory-augmented methods, the framework improves performance across multiple reasoning settings and reduces redundant LLM interactions through utility-aware memory regulation. These findings suggest that structured memory organization with importance-aware selection provides an effective balance between reasoning capability and long-term memory efficiency.

\clearpage

\bibliographystyle{apalike}  
\bibliography{references}    

@article{xi2025rise,
  author = {Xi, Zhiheng and Chen, Wenxiang and Guo, Xin and He, Wei and Ding, Yiwen and Hong, Boyang and Zhang, Ming and Wang, Junzhe and Jin, Senjie and Zhou, Enyu and Zheng, Rui and Fan, Xiaoran and Wang, Xiao and Xiong, Limao and Zhou, Yuhao and Wang, Weiran and Jiang, Changhao and Zou, Yicheng and Li, Xiangyang and Zhang, Chao and Yin, Zhangyue and Xu, Shichao and Tang, Qiong and Wang, Ziyue and Wu, Yan and Guo, Zhipeng and Zhao, Jun and Shao, Ruivuan and Li, Zejun and Li, Xuefeng and Lin, Dahua and Fujitake, Kuroiwa and Zhang, Zheng and Shao, Jing and Gui, Tao and Zhang, Qi},
  title = {The Rise and Potential of {Large Language Model}-Based Agents: A Survey},
  journal = {Science China Information Sciences},
  volume = {68},
  number = {1},
  pages = {121101},
  year = {2025}
}

@article{zhang2025survey,
  title={A survey on the memory mechanism of large language model-based agents},
  author={Zhang, Zeyu and Dai, Quanyu and Bo, Xiaohe and Ma, Chen and Li, Rui and Chen, Xu and Zhu, Jieming and Dong, Zhenhua and Wen, Ji-Rong},
  journal={ACM Transactions on Information Systems},
  volume={43},
  number={6},
  pages={1--47},
  year={2025}
}

@article{xu2025a,
  title={A-mem: Agentic memory for llm agents},
  author={Xu, Wujiang and Liang, Zujie and Mei, Kai and Gao, Hang and Tan, Juntao and Zhang, Yongfeng},
  journal={arXiv preprint arXiv:2502.12110},
  year={2025}
}

@inproceedings{zhong2024memorybank,
  author = {Zhong, Wanjun and Guo, Lianghong and Gao, Qiqi and Ye, He and Wang, Yanlin},
  title = {{MemoryBank}: Enhancing Large Language Models with Long-Term Memory},
  booktitle = {Proceedings of the AAAI Conference on Artificial Intelligence},
  volume = {38},
  pages = {19724--19731},
  year = {2024}
}

@inproceedings{kang2025memoryos,
  title={Memory os of ai agent},
  author={Kang, Jiazheng and Ji, Mingming and Zhao, Zhe and Bai, Ting},
  booktitle={Proceedings of the 2025 Conference on Empirical Methods in Natural Language Processing},
  pages={25972--25981},
  year={2025}
}

@article{fang2025lightmem,
  title={Lightmem: Lightweight and efficient memory-augmented generation},
  author={Fang, Jizhan and Deng, Xinle and Xu, Haoming and Jiang, Ziyan and Tang, Yuqi and Xu, Ziwen and Deng, Shumin and Yao, Yunzhi and Wang, Mengru and Qiao, Shuofei and others},
  journal={arXiv preprint arXiv:2510.18866},
  year={2025}
}

@inproceedings{salama2025meminsight,
  title={Meminsight: Autonomous memory augmentation for llm agents},
  author={Salama, Rana and Cai, Jason and Yuan, Michelle and Currey, Anna and Sunkara, Monica and Zhang, Yi and Benajiba, Yassine},
  booktitle={Proceedings of the 2025 Conference on Empirical Methods in Natural Language Processing},
  pages={33124--33140},
  year={2025}
}

@article{yu2026agentic,
  title={Agentic memory: Learning unified long-term and short-term memory management for large language model agents},
  author={Yu, Yi and Yao, Liuyi and Xie, Yuexiang and Tan, Qingquan and Feng, Jiaqi and Li, Yaliang and Wu, Libing},
  journal={arXiv preprint arXiv:2601.01885},
  year={2026}
}

@inproceedings{maharana2024locomo,
  title={Evaluating very long-term conversational memory of llm agents},
  author={Maharana, Adyasha and Lee, Dong-Ho and Tulyakov, Sergey and Bansal, Mohit and Barbieri, Francesco and Fang, Yuwei},
  booktitle={Proceedings of the 62nd Annual Meeting of the Association for Computational Linguistics (Volume 1: Long Papers)},
  pages={13851--13870},
  year={2024}
}

@inproceedings{wu2025longmemeval,
  title={LongMemEval: Benchmarking Chat Assistants on Long-Term Interactive Memory},
  author={Di Wu and Hongwei Wang and Wenhao Yu and Yuwei Zhang and Kai-Wei Chang and Dong Yu},
  booktitle={International Conference on Learning Representations (ICLR)},
  year={2025}
}

@article{liu2023tim,
  title={Think-in-memory: Recalling and post-thinking enable llms with long-term memory},
  author={Liu, Lei and Yang, Xiaoyan and Shen, Yue and Hu, Binbin and Zhang, Zhiqiang and Gu, Jinjie and Zhang, Guannan},
  journal={arXiv preprint arXiv:2311.08719},
  year={2023}
}

@article{packer2023memgpt,
  author = {Packer, Charles and Fang, Vivian and Patil, Shishir G. and Lin, Kevin and Wooders, Sarah and Gonzalez, Joseph E.},
  title = {{MemGPT}: Towards {LLM}s as Operating Systems},
  journal = {arXiv preprint},
  eprint = {2310.08560},
  year = {2023}
}

@article{wang2023scm,
  author = {Wang, Bing and Liang, Xinnian and Yang, Jian and Huang, Hui and Wu, Shuangzhi and Wu, Peihao and Lu, Lu and Ma, Zejun and Li, Zhoujun},
  title = {Enhancing {Large Language Model} with Self-Controlled Memory Framework},
  journal = {arXiv preprint},
  eprint = {2304.13343},
  year = {2023}
}

@article{chhikara2025mem0,
  title={Mem0: Building production-ready ai agents with scalable long-term memory},
  author={Chhikara, Prateek and Khant, Dev and Aryan, Saket and Singh, Taranjeet and Yadav, Deshraj},
  journal={arXiv preprint arXiv:2504.19413},
  year={2025}
}

@article{ocker2025grounded,
  title={A grounded memory system for smart personal assistants},
  author={Ocker, Felix and Deigm{\"o}ller, J{\"o}rg and Smirnov, Pavel and Eggert, Julian},
  journal={arXiv preprint arXiv:2505.06328},
  year={2025}
}

@article{behrouz2025titans,
  title={Titans: Learning to memorize at test time},
  author={Behrouz, Ali and Zhong, Peilin and Mirrokni, Vahab},
  journal={arXiv preprint arXiv:2501.00663},
  year={2024}
}

@inproceedings{Du2025RethinkingMI,
  author = {Du, Yiming and Huang, Wenyu and Zheng, Danna and Wang, Zhaowei and Montella, S{\'e}bastien and Lapata, Mirella and Wong, Kam-Fai and Pan, Jeff Z.},
  title = {Rethinking Memory in {LLM}-Based Agents: Representations, Operations, and Emerging Topics},
  booktitle = {Proceedings of the 2025 Conference on Empirical Methods in Natural Language Processing},
  year = {2025}
}

@article{Mei2024AIOSLA,
  title={Aios: Llm agent operating system},
  author={Mei, Kai and Zhu, Xi and Xu, Wujiang and Hua, Wenyue and Jin, Mingyu and Li, Zelong and Xu, Shuyuan and Ye, Ruosong and Ge, Yingqiang and Zhang, Yongfeng},
  journal={arXiv preprint arXiv:2403.16971},
  year={2024}
}

@article{Modarressi2023RETLLMTA,
  title={Ret-llm: Towards a general read-write memory for large language models},
  author={Modarressi, Ali and Imani, Ayyoob and Fayyaz, Mohsen and Sch{\"u}tze, Hinrich},
  journal={arXiv preprint arXiv:2305.14322},
  year={2023}
}

@inproceedings{papineni-etal-2002-bleu,
  title={Bleu: a method for automatic evaluation of machine translation},
  author={Papineni, Kishore and Roukos, Salim and Ward, Todd and Zhu, Wei-Jing},
  booktitle={Proceedings of the 40th annual meeting of the Association for Computational Linguistics},
  pages={311--318},
  year={2002}
}

@inproceedings{Pan2025OnMC,
  author = {Pan, Zhuoshi and Wu, Qianhui and Jiang, Huiqiang and Luo, Xufang and Cheng, Hao and Li, Dongsheng and Yang, Yuqing and Lin, Chin-Yew and Zhao, H. Vicky and Qiu, Lili and Gao, Jianfeng},
  title = {SeCom: On Memory Construction and Retrieval for Personalized Conversational Agents},
  booktitle = {International Conference on Learning Representations},
  year = {2025}
}

@article{rasmussen2025zep,
  title={Zep: a temporal knowledge graph architecture for agent memory},
  author={Rasmussen, Preston and Paliychuk, Pavlo and Beauvais, Travis and Ryan, Jack and Chalef, Daniel},
  journal={arXiv preprint arXiv:2501.13956},
  year={2025}
}

@article{edge2024graphrag,
  title={From local to global: A graph rag approach to query-focused summarization},
  author={Edge, Darren and Trinh, Ha and Cheng, Newman and Bradley, Joshua and Chao, Alex and Mody, Apurva and Truitt, Steven and Metropolitansky, Dasha and Ness, Robert Osazuwa and Larson, Jonathan},
  journal={arXiv preprint arXiv:2404.16130},
  year={2024}
}

@inproceedings{jiang2023flare,
  title={Active retrieval augmented generation},
  author={Jiang, Zhengbao and Xu, Frank F and Gao, Luyu and Sun, Zhiqing and Liu, Qian and Dwivedi-Yu, Jane and Yang, Yiming and Callan, Jamie and Neubig, Graham},
  booktitle={Proceedings of the 2023 conference on empirical methods in natural language processing},
  pages={7969--7992},
  year={2023}
}

@inproceedings{trivedi2023ircot,
  title={Interleaving retrieval with chain-of-thought reasoning for knowledge-intensive multi-step questions},
  author={Trivedi, Harsh and Balasubramanian, Niranjan and Khot, Tushar and Sabharwal, Ashish},
  booktitle={Proceedings of the 61st annual meeting of the association for computational linguistics (volume 1: long papers)},
  pages={10014--10037},
  year={2023}
}

@inproceedings{lin2024radit,
  author = {Lin, Xi Victoria and Chen, Xilun and Chen, Mingda and Shi, Weijia and Lomeli, Maria and James, Rich and Rodriguez, Pedro and Kahn, Jacob and Szilvasy, Gergely and Lewis, Mike and Zettlemoyer, Luke and Yih, Scott},
  title = {RA-DIT: Retrieval-Augmented Dual Instruction Tuning},
  booktitle = {International Conference on Learning Representations},
  year = {2024}
}

@inproceedings{tang2026activerag,
  author = {Tang, Tao and Yue, Xiaodong and Chen, Yufei and Shi, Jie and Ding, Shijie},
  title = {Active Retrieval-Augmented Generation with Conflict-Fused Uncertainty Quantification},
  booktitle = {Proceedings of the ACM Web Conference},
  pages = {8629--8632},
  year = {2026}
}

@inproceedings{yu2024chainofnote,
  author = {Yu, Wenhao and Zhang, Hongming and Pan, Xiaoman and Cao, Peixin and Ma, Kaixin and Li, Jian and Wang, Hongwei and Yu, Dong},
  title = {Chain-of-Note: Enhancing Robustness in Retrieval-Augmented Language Models},
  booktitle = {Proceedings of the 2024 Conference on Empirical Methods in Natural Language Processing},
  pages = {14672--14685},
  year = {2024}
}

@inproceedings{wang2025scm,
  author={Wang, Bing and Liang, Xinnian and Yang, Jian and Huang, Hui and Wu, Zhenhe and Wu, ShuangZhi and Ma, Zejun and Li, Zhoujun},
  title={Scm: Enhancing large language model with self-controlled memory framework},
  booktitle={International Conference on Database Systems for Advanced Applications},
  pages={188--203},
  year={2025}
}

@article{liu2024agentlite,
  author = {Liu, Zhiwei and Yao, Weiran and Zhang, Jianguo and Yang, Liangwei and Liu, Zuxin and Tan, Juntao and Choubey, Prafulla K. and Lan, Tian and Wu, Jason and Wang, Huan and Heinecke, Shelby and Xiong, Caiming and Savarese, Silvio},
  title = {AgentLite: A Lightweight Library for Building and Advancing Task-Oriented LLM Agent Systems},
  journal = {arXiv preprint},
  eprint = {2402.15538},
  year = {2024}
}

@inproceedings{wang2025openhands,
  author = {Wang, Xingyao and Li, Boxuan and Song, Yufan and Xu, Frank F. and Tang, Xiangru and Zhuge, Mingchen and Pan, Jiayi and Song, Yueqi and Li, Bowen and Singh, Jaskirat and Tran, Hoang and Li, Fuqiang and Ma, Ren and Zheng, Mingzhang and Qian, Bill and Shao, Daniel and Muennighoff, Niklas and Zhang, Yizhe and Hui, Binyuan and Lin, Junyang and Brennan, Robert and Peng, Hao and Ji, Heng and Neubig, Graham},
  title = {OpenHands: An Open Platform for AI Software Developers as Generalist Agents},
  booktitle = {International Conference on Learning Representations},
  year = {2025}
}

@inproceedings{li2025hello,
  author={Li, Hao and Yang, Chenghao and Zhang, An and Deng, Yang and Wang, Xiang and Chua, Tat-Seng},
  title={Hello again! llm-powered personalized agent for long-term dialogue},
  booktitle={Proceedings of the 2025 Conference of the Nations of the Americas Chapter of the Association for Computational Linguistics: Human Language Technologies (Volume 1: Long Papers)},
  pages={5259--5276},
  year={2025}
}

@article{liu2023lostmiddle,
  author={Liu, Nelson F and Lin, Kevin and Hewitt, John and Paranjape, Ashwin and Bevilacqua, Michele and Petroni, Fabio and Liang, Percy},
  title={Lost in the middle: How language models use long contexts},
  journal={Transactions of the association for computational linguistics},
  volume={12},
  pages={157--173},
  year={2024}
}

@inproceedings{lewis2020rag,
  author = {Lewis, Patrick and Perez, Ethan and Piktus, Aleksandra and Petroni, Fabio and Karpukhin, Vladimir and Goyal, Naman and K{\"u}ttler, Heinrich and Lewis, Mike and Yih, Wen-tau and Rockt{\"a}schel, Tim and Riedel, Sebastian and Kiela, Douwe},
  title = {Retrieval-Augmented Generation for Knowledge-Intensive NLP Tasks},
  booktitle = {Advances in Neural Information Processing Systems 33 (NeurIPS 2020)},
  year = {2020}
}

@article{zhang2025ascot,
  author={Zhang, Dongxu and Yang, Ning and Zhu, Jihua and Yang, Jinnan and Xin, Miao and Tian, Baoliang},
  title={Ascot: An adaptive self-correction chain-of-thought method for late-stage fragility in llms},
  journal={arXiv preprint arXiv:2508.05282},
  year={2025}
}

@article{zhang2026not,
  title={Not all queries need deep thought: Coficot for adaptive coarse-to-fine stateful refinement},
  author={Zhang, Dongxu and Lin, Hongqiang and Sun, Yiding and Wang, Pengyu and Wang, Qirui and Yang, Ning and Zhu, Jihua},
  journal={arXiv preprint arXiv:2603.08251},
  year={2026}
}

@article{zhang2026chain,
  title={Chain-of-thought compression should not be blind: V-skip for efficient multimodal reasoning via dual-path anchoring},
  author={Zhang, Dongxu and Sun, Yiding and Tan, Cheng and Yan, Wenbiao and Yang, Ning and Zhu, Jihua and Zhang, Haijun},
  journal={arXiv preprint arXiv:2601.13879},
  year={2026}
}

\appendix

\section{Formal Theoretical Proofs}

This appendix presents the formal analytical derivation of the theoretical claims introduced in the main text. The proofs are structured to ensure completeness and self-consistency, adhering to the rigorous evaluation standards for agentic memory dynamics.

\subsection{Proof of Theorem 1: Hierarchical Retrieval Priority}\label{appendix:theorem1}

\begin{customthm}{1}[Restatement of Theorem \ref{theorem1}]
Under the structural gap condition $\gamma > 1 + \delta$, core factual invariants in $L_1$ are strictly ranked above distractors in $L_4$ under Eq.~\ref{eq:5}, ensuring stable Top-$K$ retrieval.
\end{customthm}

\paragraph{Intuition.} 
The core idea is that the hierarchical weighting system creates a "score buffer" that is sufficiently large to counteract any semantic noise (similarities that happen to be high for irrelevant information).

Let $m_{core} \in L_1$ be a factual invariant and $m_{noise} \in L_4$ be a semantic distractor. The retrieval scoring function defined in Section 3.5 is:
\begin{equation}
    R(m, q) = \omega(L) \cdot \text{sim}(v_m, v_q).
\end{equation}

Consider the score ratio between the core memory and the noise:
\begin{equation}
    \rho = \frac{R(m_{core}, q)}{R(m_{noise}, q)} = \frac{\omega(L_1)}{\omega(L_4)} \cdot \frac{\text{sim}(v_{core}, v_q)}{\text{sim}(v_{noise}, v_q)}.
\end{equation}

From \textbf{Assumption A (Structural Gap)}, we have $\frac{\omega(L_1)}{\omega(L_4)} \geq \gamma$. From our \textbf{Noise-Bounded} assumption, we know that the similarity of noise is bounded by a factor of $(1+\delta)$ relative to the core fact, i.e., $\frac{\text{sim}(v_{core}, v_q)}{\text{sim}(v_{noise}, v_q)} \geq \frac{1}{1+\delta}$. Substituting these into the ratio:
\begin{equation}
    \rho \geq \frac{\gamma}{1+\delta}.
\end{equation}

If the system hyper-parameters are tuned such that $\gamma > 1+\delta$, then $\rho > 1$, which implies $R(m_{core}, q) > R(m_{noise}, q)$. Since SF-AMS employs a Top-$K$ retrieval strategy over the combined candidate set, $m_{core}$ will always be strictly higher than $m_{noise}$ in the ranking queue. This ensures that core invariants are never displaced by noise distractors of high-similarity in the context window.

\subsection{Proof of Proposition 1: Stability of Memory Dynamics}
\label{appendix:A2}

\begin{customthm}{1}[Restatement of Proposition \ref{proposition1}]
Under the memory dynamics defined in Eq.~\ref{eq:3}, and assuming bounded reinforcement and capacity-constrained updates, the aggregate survival potential $V(X_t)$ remains uniformly bounded over time.
\end{customthm}

\paragraph{State Definition.}
At time $t$, the system state is defined as $X_t = (M_t, \Phi_t)$, where $M_t$ denotes the set of active memory units with $|M_t| \le N_{\max}$, and $\Phi_t = \{\Phi_i^t\}_{i \in M_t}$ represents their survival potentials. By construction, $\Phi_i^t \ge 0$ for all $i,t$.

\paragraph{Bounded Reinforcement.}
The reinforcement term is defined as
\begin{equation}
R_i^t = I(m_i)\cdot \Gamma_{\text{usage}}^t \cdot \Psi_{\text{div}}(m_i, M_t).
\end{equation}
Since $I(m_i) \in [0,1]$, $\Gamma_{\text{usage}}^t \in \{0,1\}$, and $\Psi_{\text{div}} \in [1-\alpha_{\text{div}}, 1+\alpha_{\text{div}}]$, it follows that there exists a constant $R_{\max} > 0$ such that
\begin{equation}
0 \le R_i^t \le R_{\max}, \quad \forall i,t.
\end{equation}

\paragraph{Single-Step Dynamics.}
From the update rule,
\begin{equation}
\Phi_i^{t+1} = \max\{0, \Phi_i^t + R_i^t - \lambda\},
\end{equation}
we obtain the upper bound
\begin{equation}
\Phi_i^{t+1} \le \Phi_i^t + R_{\max}.
\end{equation}
Thus, the per-step growth of each memory unit is bounded.

\paragraph{Selective Persistence.}
The update dynamics induce a natural separation between memory units:
\begin{itemize}
    \item If $\mathbb{E}[R_i^t] > \lambda$, the unit can sustain or increase its survival potential and remain active.
    \item If $\mathbb{E}[R_i^t] < \lambda$, the unit undergoes decay and will eventually be removed.
\end{itemize}
This mechanism ensures that high-value memory units persist, while low-value or redundant ones are gradually eliminated.

\paragraph{Capacity-Constrained Dynamics.}
The system enforces a hard constraint $|M_t| \le N_{\max}$. When capacity is reached, insertion of new memory units triggers eviction of the unit with the smallest survival potential:
\begin{equation}
i^* = \arg\min_{j \in M_t \cup \{m_{\text{new}}\}} \Phi_j.
\end{equation}
This truncation prevents unbounded accumulation of memory units.

\paragraph{Bounded Aggregate Potential.}
Define the Lyapunov function:
\begin{equation}
V(X_t) = \sum_{i \in M_t} \Phi_i^t.
\end{equation}
Since $|M_t| \le N_{\max}$, we have
\begin{equation}
V(X_t) \le N_{\max} \cdot \max_{i \in M_t} \Phi_i^t.
\end{equation}

Although individual survival potentials may increase when reinforcement exceeds decay, no unit can grow without bound while remaining in the system. In particular, any unit with persistently increasing potential must continuously compete with newly inserted memory units. Due to the eviction mechanism, low-utility units are removed and excessively dominant states do not accumulate indefinitely across the memory set.

Moreover, in practical implementations, the numerical representation of $\Phi_i^t$ is inherently bounded due to finite precision and system constraints, which further prevents unbounded growth in real-world deployments. From a theoretical perspective, the continual competition induced by insertion and eviction ensures that no single unit can indefinitely dominate the memory pool.

Therefore, the aggregate potential $V(X_t)$ remains bounded over time.

\paragraph{Conclusion.}
The memory system operates in a stable regime where the number of active memory units is bounded and the aggregate survival potential does not diverge. Stability arises from the combined effects of bounded reinforcement, selective persistence, and capacity-constrained eviction, without requiring decay to dominate reinforcement.

In this sense, the system is stable in that its state remains confined within a bounded region over time.

\section{Additional Experiments and Analysis}
\subsection{Empirical Validation of Dynamics Equilibrium}
\label{app:dynamics}

This section provides empirical evidence to support the stability claims and the Foster-Lyapunov drift condition outlined in Proposition 2. The evolution of the memory state over a 100-step horizon is detailed in Table~\ref{tab:full_dynamics_data}.We vary a normalized forgetting control parameter, denoted as \textit{Ratio}, which scales the decay intensity in Eq.~\ref{eq:3}. Specifically, the effective decay is implemented as $\lambda' = \textit{Ratio} \cdot \lambda$.

\begin{table}[t]
\centering
\small

\caption{Dynamics of memory population and survival potential $\Phi$ across hierarchical layers ($L_1$--$L_4$) under varying ratio settings. 
Here, \textbf{Ratio} denotes the normalized forgetting strength, implemented as a scaling factor applied to the global decay coefficient (Eq.~\ref{eq:3}). 
Higher values correspond to more aggressive decay, which progressively shifts memory mass toward lower-priority layers while suppressing high-energy states.}
\label{tab:full_dynamics_data}

\setlength{\tabcolsep}{4pt}

\begin{tabular*}{\textwidth}{@{\extracolsep{\fill}}ccccccccc}
\toprule
\textbf{Ratio} &
\multicolumn{4}{c}{\textbf{Population (Mean Count)}} &
\multicolumn{4}{c}{\textbf{Energy (Mean $\Phi$)}} \\
\cmidrule(lr){2-5} \cmidrule(lr){6-9}
 & Core ($L_1$) & Imp. ($L_2$) & Sec. ($L_3$) & Irr. ($L_4$)
 & $L_1$ & $L_2$ & $L_3$ & $L_4$ \\
\midrule
0.00 & 35.5000 & 32.3750 & 110.3750 & 28.6250 & 0.8397 & 0.6212 & 0.4346 & 0.2396 \\
0.05 & 37.3333 & 32.6667 & 115.2667 & 29.7333 & 0.5223 & 0.3034 & 0.1564 & 0.0432 \\
0.10 & 36.5714 & 32.5000 & 119.7143 & 33.5000 & 0.1790 & 0.0733 & 0.0279 & 0.0186 \\
0.15 & 36.5714 & 32.7143 & 122.7857 & 38.6429 & 0.0607 & 0.0102 & 0.0073 & 0.0175 \\
0.20 & 37.3333 & 33.5333 & 125.2000 & 43.6000 & 0.0019 & 0.0148 & 0.0030 & 0.0180 \\
0.25 & 36.5714 & 33.8571 & 127.5714 & 49.1429 & 0.0000 & 0.0118 & 0.0064 & 0.0165 \\
0.30 & 37.3333 & 34.0000 & 131.2667 & 53.9333 & 0.0000 & 0.0012 & 0.0067 & 0.0107 \\
0.35 & 36.5714 & 34.5000 & 133.7143 & 59.1429 & 0.0000 & 0.0078 & 0.0071 & 0.0113 \\
0.40 & 36.5714 & 35.0714 & 136.2143 & 63.6429 & 0.0000 & 0.0112 & 0.0063 & 0.0100 \\
0.45 & 37.3333 & 35.3333 & 141.7333 & 67.2000 & 0.0000 & 0.0073 & 0.0099 & 0.0085 \\
0.50 & 36.5714 & 35.2143 & 144.6429 & 71.7857 & 0.0000 & 0.0023 & 0.0086 & 0.0082 \\
0.55 & 37.3333 & 35.6000 & 150.6667 & 75.0000 & 0.0000 & 0.0053 & 0.0102 & 0.0045 \\
0.60 & 36.5714 & 35.7857 & 154.6429 & 77.7143 & 0.0000 & 0.0087 & 0.0108 & 0.0044 \\
0.65 & 36.5714 & 35.9286 & 158.0714 & 82.2857 & 0.0000 & 0.0066 & 0.0076 & 0.0068 \\
0.70 & 37.3333 & 36.0000 & 163.3333 & 86.3333 & 0.0000 & 0.0020 & 0.0066 & 0.0062 \\
0.75 & 36.5714 & 35.9286 & 165.2143 & 91.5714 & 0.0000 & 0.0001 & 0.0047 & 0.0092 \\
0.80 & 37.3333 & 36.1333 & 169.1333 & 97.0667 & 0.0000 & 0.0026 & 0.0050 & 0.0071 \\
0.85 & 36.5714 & 36.2857 & 171.7143 & 101.1429 & 0.0000 & 0.0053 & 0.0076 & 0.0041 \\
0.90 & 36.5714 & 36.2857 & 174.2143 & 106.6429 & 0.0000 & 0.0028 & 0.0049 & 0.0067 \\
0.95 & 37.3333 & 36.3333 & 178.1333 & 113.0667 & 0.0000 & 0.0004 & 0.0046 & 0.0058 \\
1.00 & 36.0000 & 36.3750 & 177.3750 & 116.6250 & 0.0000 & 0.0022 & 0.0034 & 0.0048 \\
\bottomrule
\end{tabular*}
\end{table}

\begin{figure}[htbp]
    \centering
    \begin{minipage}{0.48\textwidth}
        \centering
        \includegraphics[width=\textwidth]{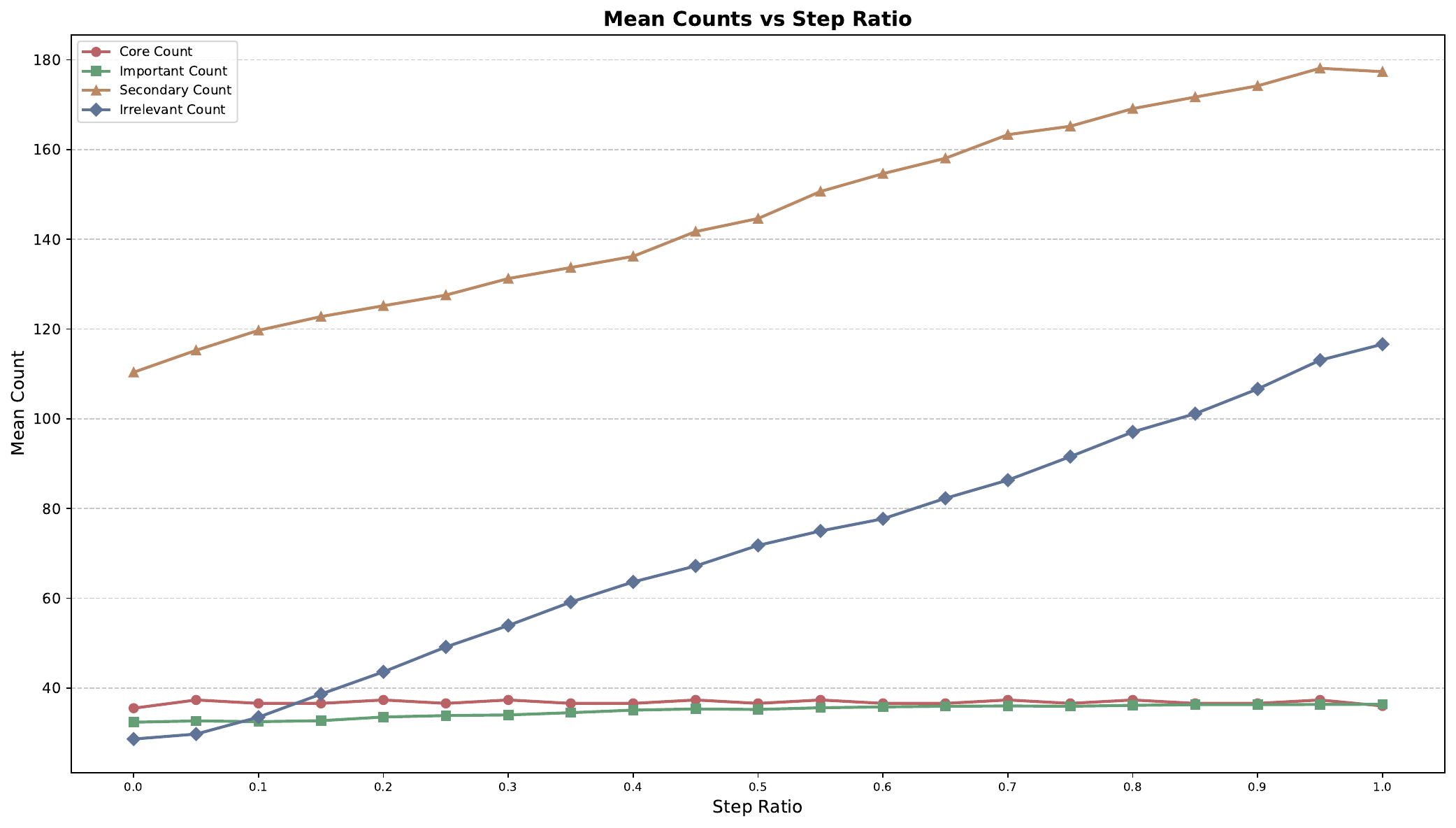}
        \caption{Evolution of memory layer populations. The system reaches a dynamic equilibrium where core facts are preserved while redundant noise is cyclically purged.}
        \label{fig:layer_counts}
    \end{minipage}
    \hfill
    \begin{minipage}{0.48\textwidth}
        \centering
        \includegraphics[width=\textwidth]{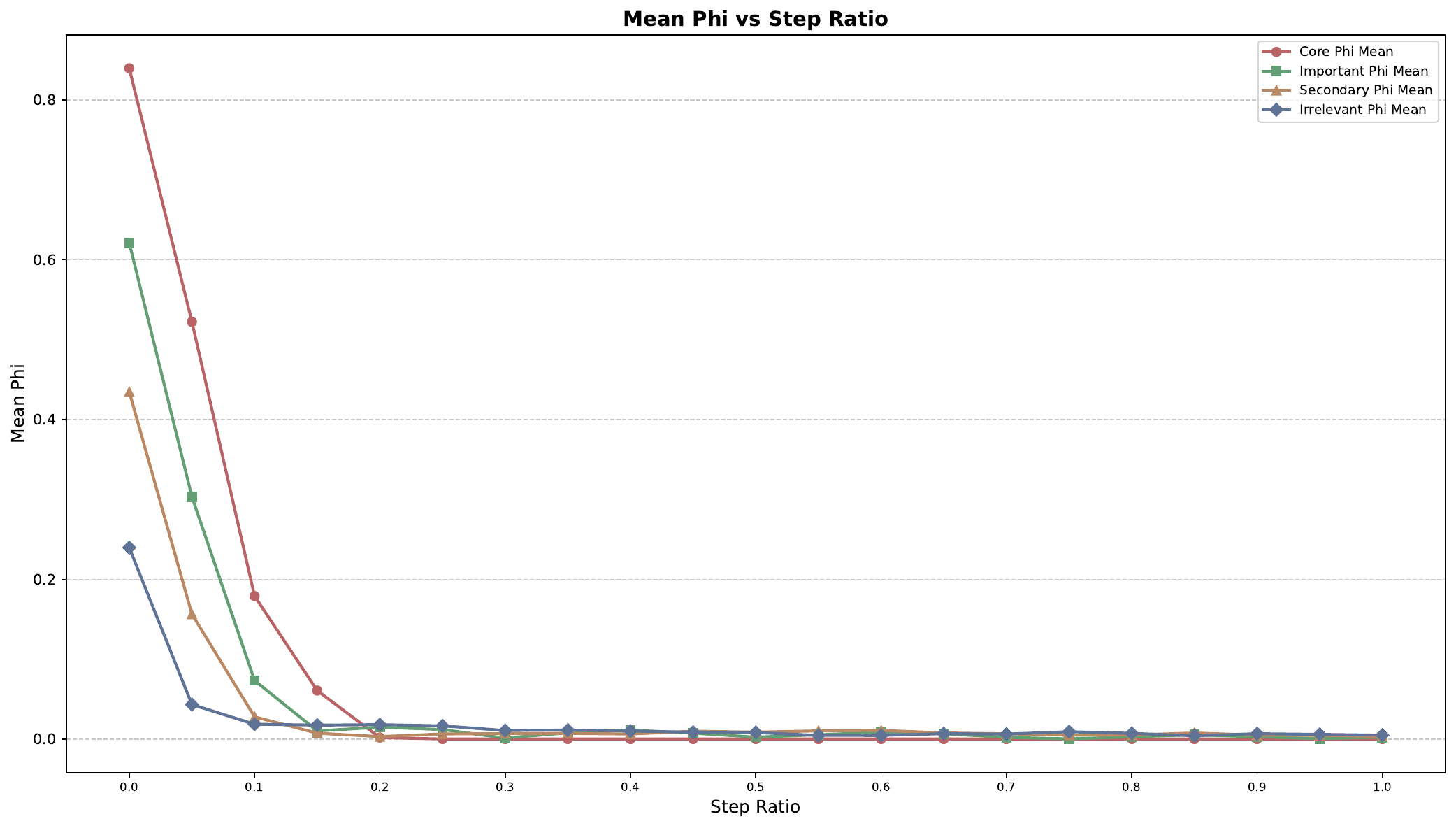}
        \caption{Mean survival potential $\Phi$ across different layers. The rapid initial decay followed by a stabilized gap confirms the effectiveness of the decay-driven forgetting mechanism.}
        \label{fig:mean_phi}
    \end{minipage}
\end{figure}

As illustrated by the survival potential $\Phi$ trajectories (Table~\ref{tab:full_dynamics_data} and Figure~\ref{fig:mean_phi}), the system undergoes a rapid decay toward a low-energy regime during the initial interaction phase. This numerical trend offers two critical insights:

\begin{itemize}
    \item \textbf{Transition to a ``Ready-to-Evict'' State:} The convergence of $\Phi$ toward a near-zero baseline in the later stages does not indicate information loss. Instead, it signifies that the memory has entered a stable, low-energy state. By maintaining this equilibrium, SF-AMS intentionally maximizes the ``survival space'' for high-entropy novel information, ensuring the memory remains receptive without breaching capacity limits.
    
    \item \textbf{Mechanism of Efficiency Gains:} This energy stratification provides the physical basis for the 10$\times$ speedup observed in our main results. With a clear potential gap established ($\Phi_{L_1} \gg \Phi_{L_4}$), the retrieval module can near-instantaneously bypass low-energy noise. This allows computational resources to be focused exclusively on the core skeletal memory, significantly reducing the search space compared to unstructured baselines.
\end{itemize} 

\subsection{Memory Capacity Stress Test}

\begin{figure}[H]
    \centering
    \includegraphics[width=\textwidth]{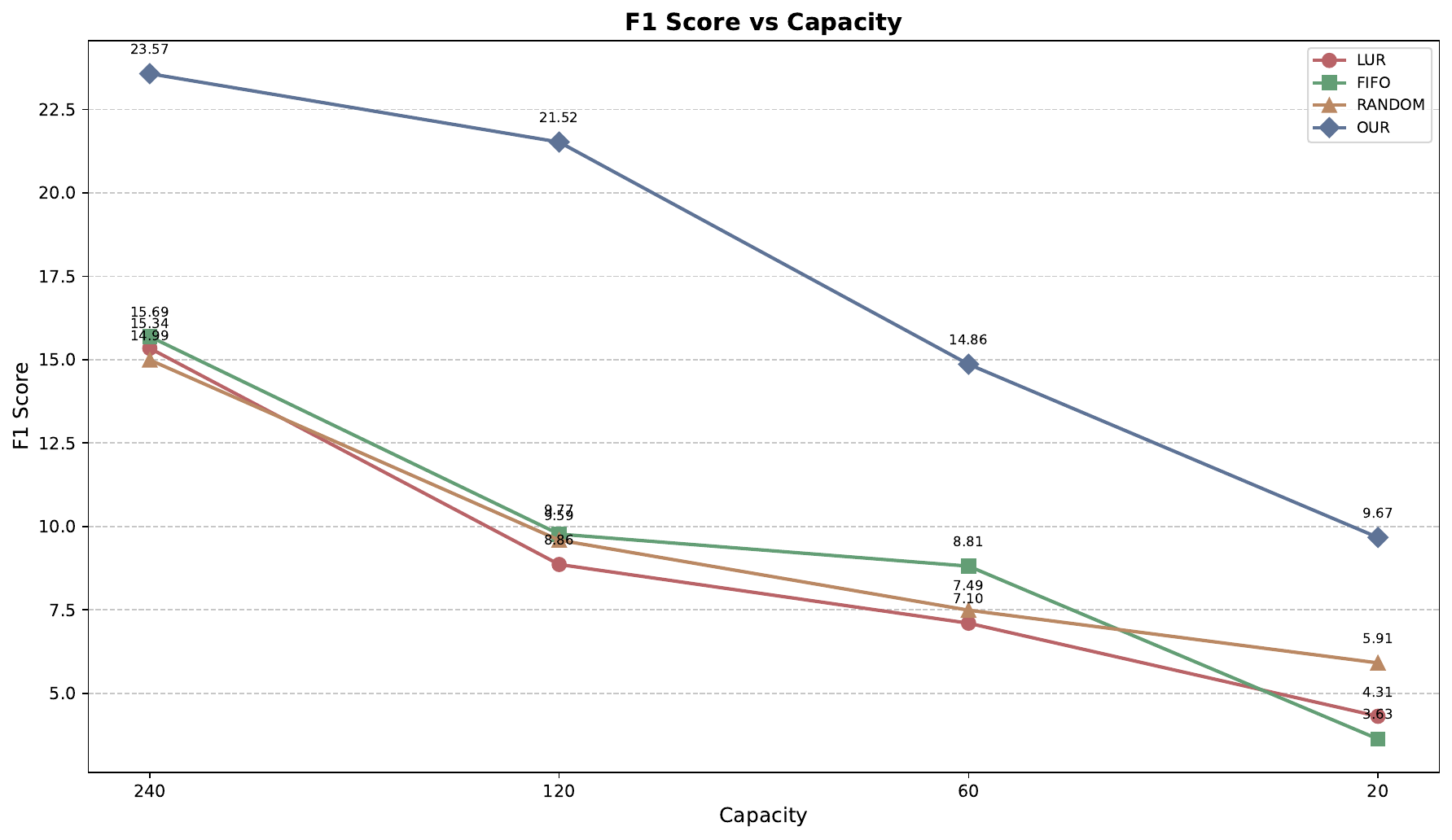} 
    \caption{Performance under varying global memory capacity $N_{\max}$. SF-AMS consistently outperforms LRU, FIFO, and RANDOM baselines across all settings and exhibits strong robustness under low-capacity conditions.}
    \label{fig:capacity_stress}
\end{figure}

To evaluate robustness under extreme resource constraints, we vary the maximum capacity $N_{\max} \in \{20, 60, 120, 240\}$. This directly tests the operational limits of the system under different memory budgets.

As shown in Figure~\ref{fig:capacity_stress}, while all methods benefit from increased memory budgets, heuristic baselines like FIFO and LRU exhibit irregular scaling and diminishing returns. In contrast, SF-AMS maintains the highest F1 scores across all capacity levels, achieving a notable 9.67 F1 even under the most restrictive setting ($N=20$). This monotonic improvement trend suggests that our priority-driven retention is more efficient at utilizing marginal increases in memory than simple recency or random-based strategies.

\begin{table}[t]
\centering
\small

\caption{Performance under different types of retrieval distractors. SF-AMS consistently outperforms the vanilla baseline across all categories, demonstrating strong robustness to noise.}
\label{tab:noise_types}

\setlength{\tabcolsep}{6pt}
\renewcommand{\arraystretch}{1.2}

\begin{tabular*}{\textwidth}{@{\extracolsep{\fill}}lcccc}
\toprule
\textbf{Category} & \textbf{Count} & \textbf{Metric} & \textbf{Vanilla} & \textbf{SF-AMS} \\
\midrule
\textbf{Event}     & 983 & \textbf{Acc@1}       & 16.67   & 86.67 \\
\textbf{Temporal}  & 201 & \textbf{Rec.Rate}   & -       & 70.00 \\
\textbf{Nearby}    & 316 & \textbf{Margin}      & -0.12 & 0.13 \\
\bottomrule
\end{tabular*}
\end{table}

\begin{figure}[t]
\centering
\includegraphics[width=\textwidth]{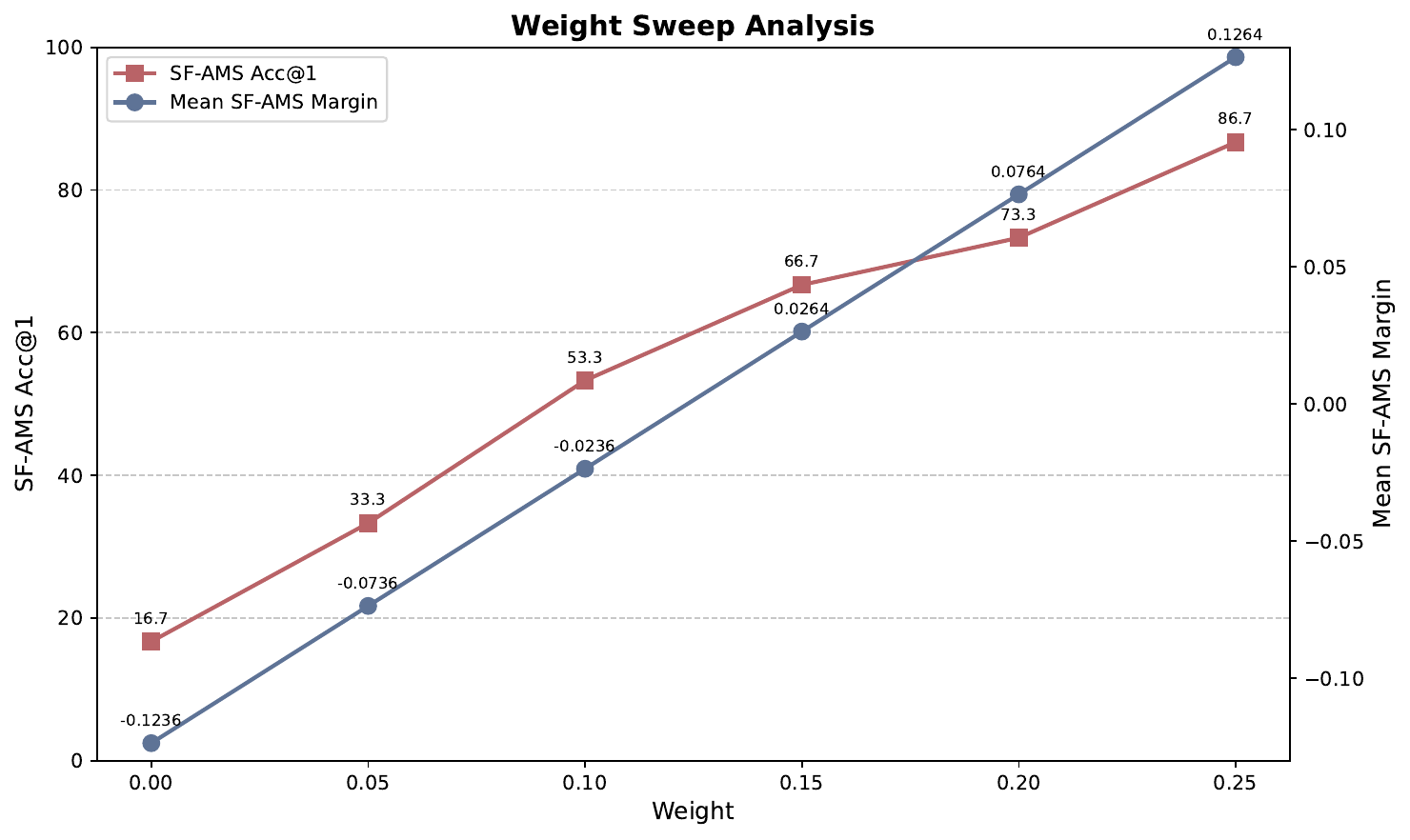}
\caption{Sensitivity of retrieval performance with respect to $\gamma$. The model shows stable performance across a wide range of $\gamma$, indicating low hyperparameter sensitivity.}
\label{fig:gamma_sensitivity}
\end{figure}

To assess retrieval stability under noisy conditions, we introduce naturally occurring ``hard negatives'' from the LoCoMo dataset—distractors that share high semantic similarity with the query but differ in factual or temporal correctness. 

The results in Table~\ref{tab:noise_types} indicate that vanilla dense retrieval is highly susceptible to temporal and event confusions, often selecting misleading but high-similarity memories. Conversely, SF-AMS achieves a significant accuracy boost (86.67\% Acc@1) and maintains a positive ranking margin (0.1264). This robustness stems from the combination of keyword-level anchoring and memory prioritization, which effectively filters out distractors that lack structural importance. Furthermore, Figure~\ref{fig:gamma_sensitivity} demonstrates that the system remains stable across the hierarchical weight $\gamma$ range, with improved separation as the structural signals are prioritized.

\section{Prompt Templates and Examples}\label{appendix:prompts}
\subsection{Generate System Response System Prompt}
\begin{figure}[H]
    \centering
    \includegraphics[width=0.85\textwidth]{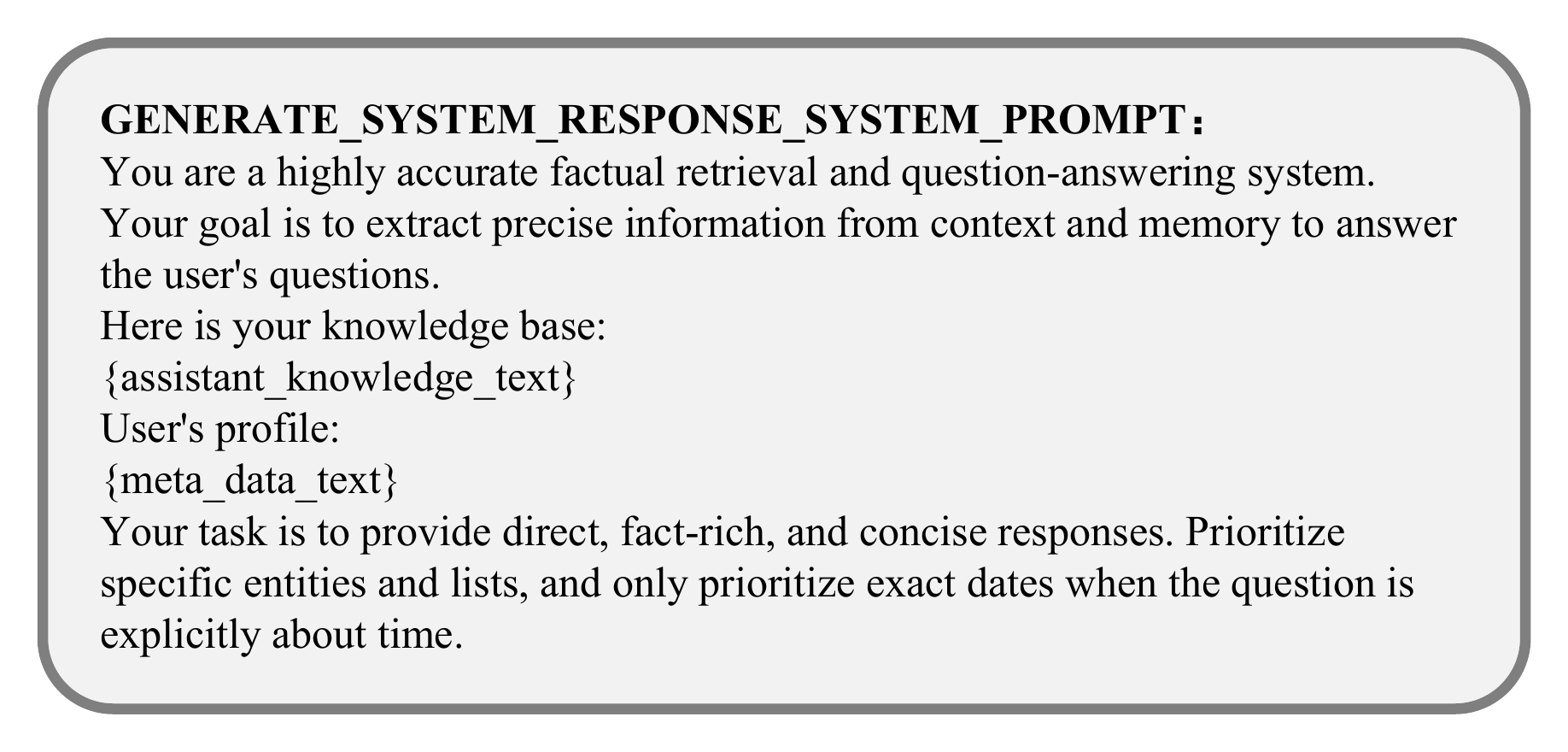}
    \label{fig:prompt1}
\end{figure}

\subsection{Generate System Response User Prompt}
\begin{figure}[H]
    \centering
    \includegraphics[width=0.85\textwidth]{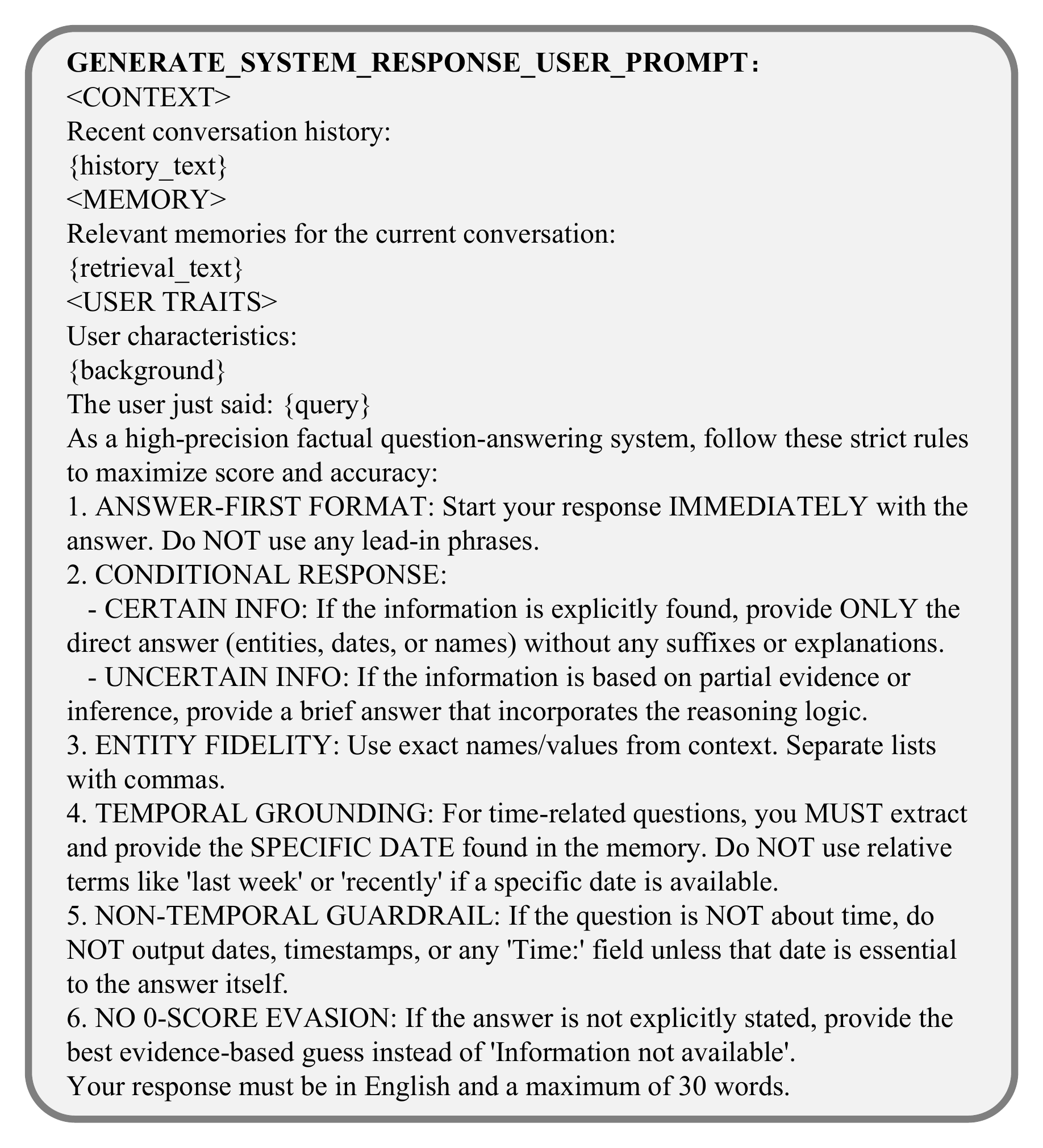}
    \label{fig:prompts2}
\end{figure}

\subsection{Evaluate Memory Importance}
\begin{figure}[H]
    \centering
    \includegraphics[width=0.85\textwidth]{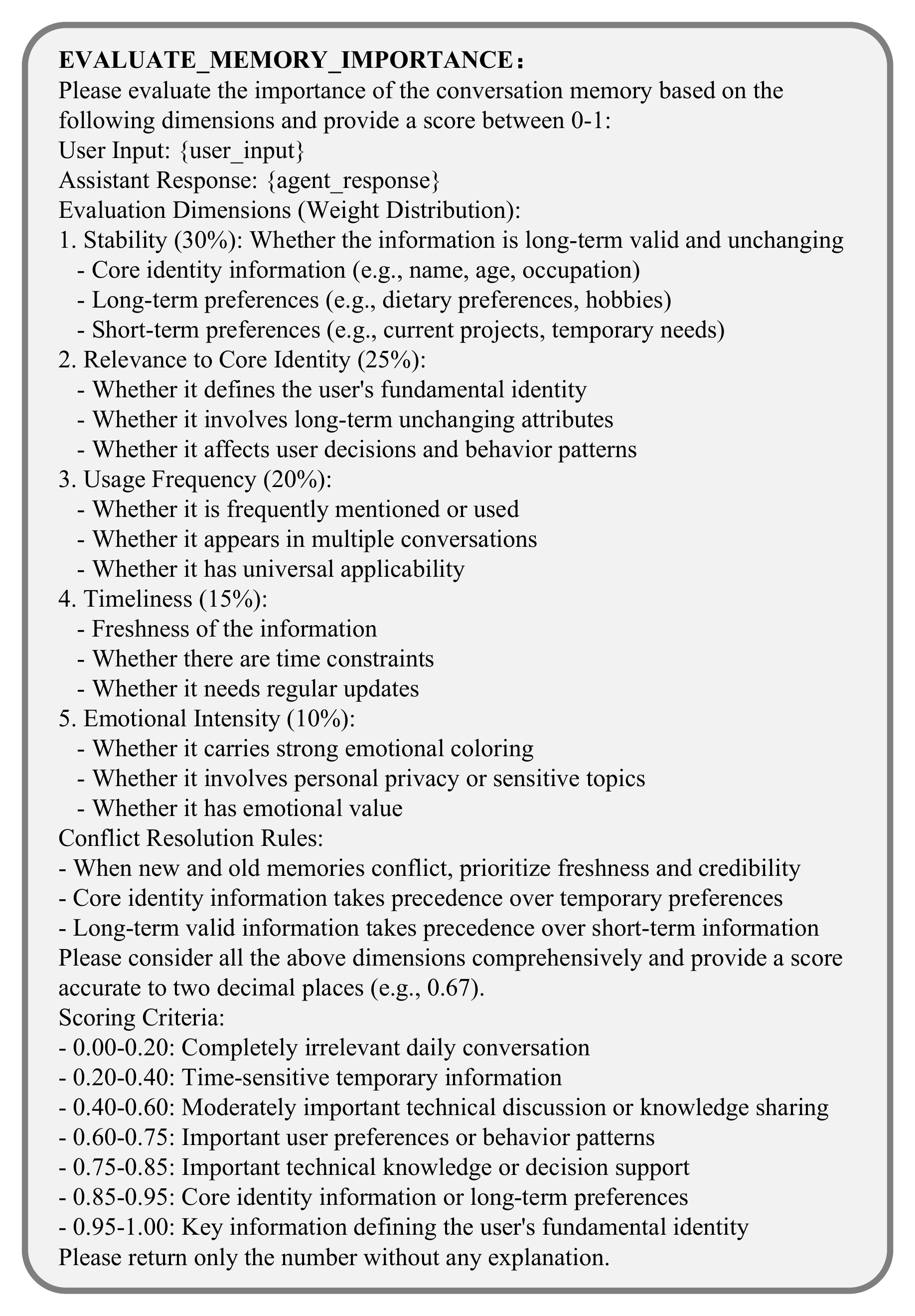}
    \label{fig:prompts3}
\end{figure}

\subsection{Examples of Q/A with SF-AMS}
\begin{figure}[H]
    \centering
    \includegraphics[width=0.85\textwidth]{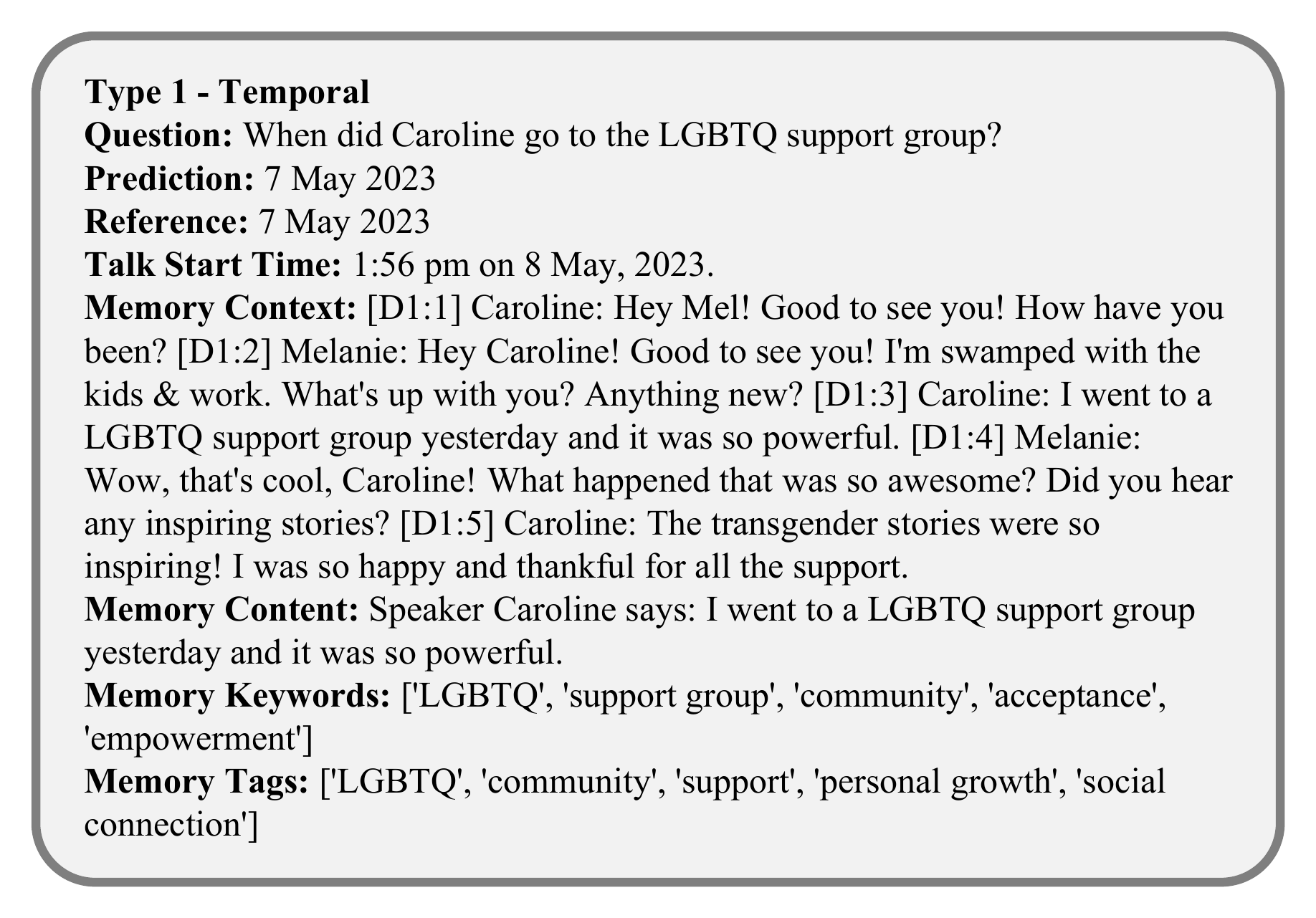}
    \label{fig:example}
\end{figure}
\begin{figure}[H]
    \centering
    \includegraphics[width=0.85\textwidth]{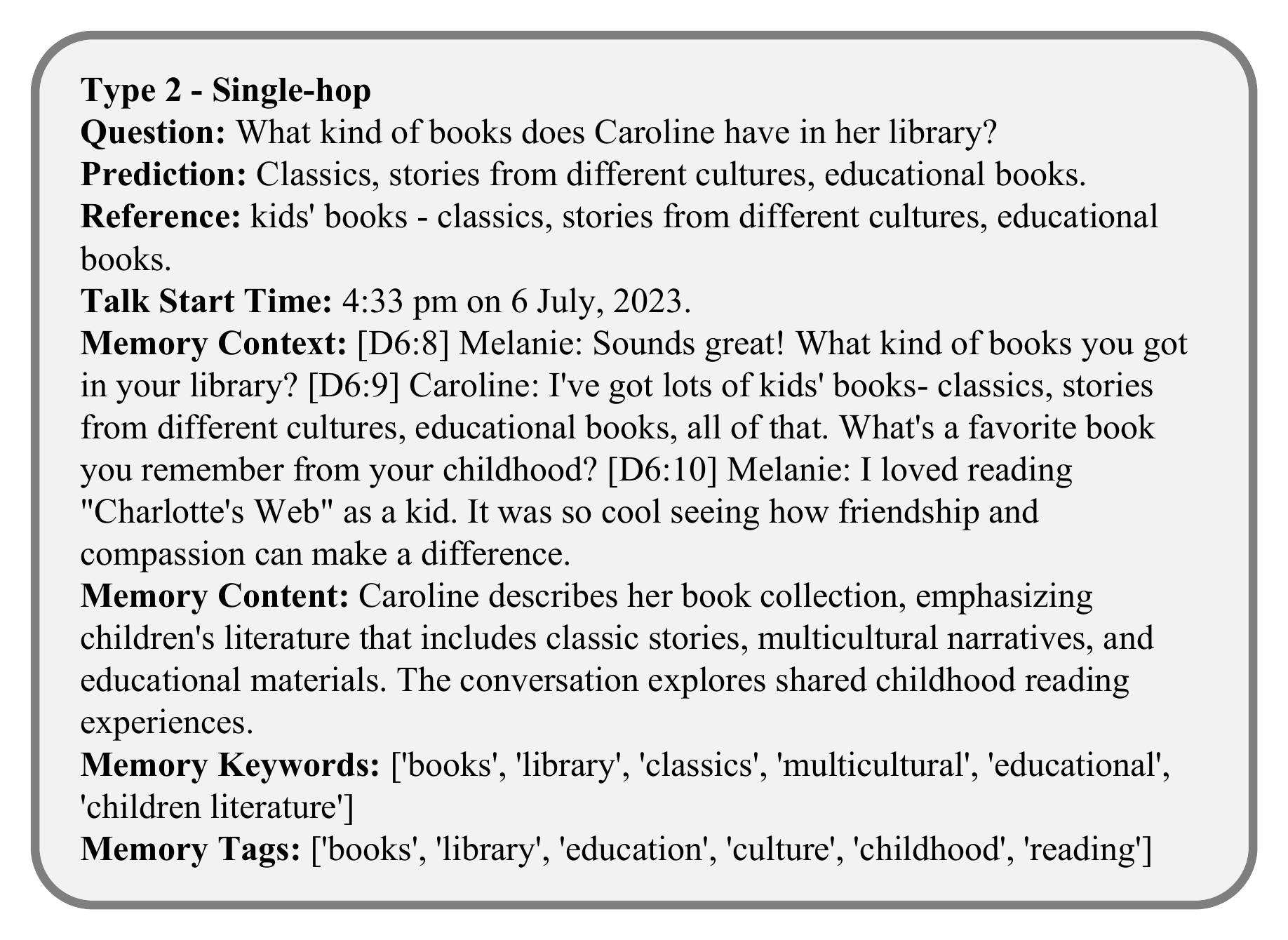}
    \label{fig:example2}
\end{figure}
\begin{figure}[H]
    \centering
    \includegraphics[width=0.85\textwidth]{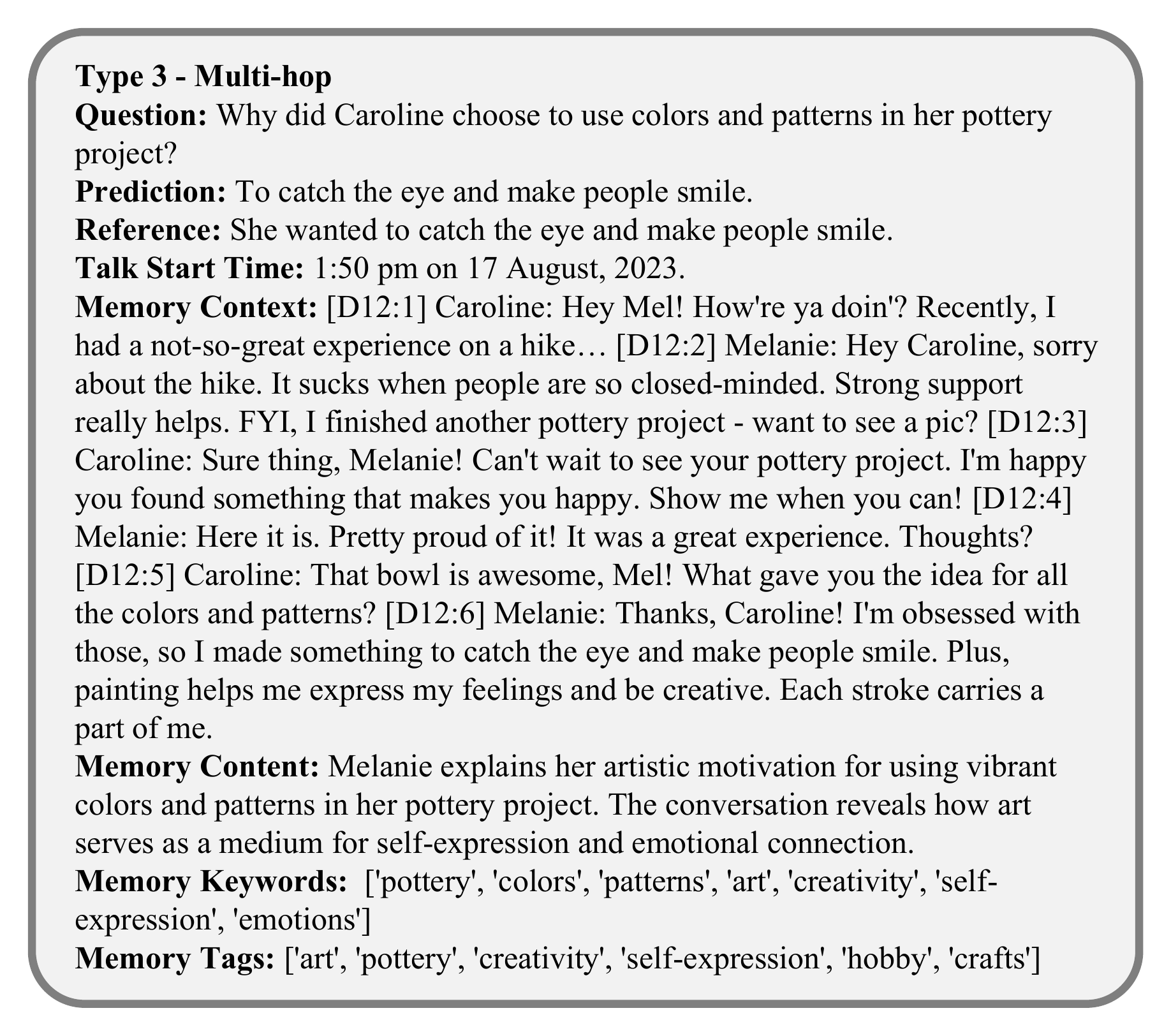}
    \label{fig:example3}
\end{figure}
\begin{figure}[H]
    \centering
    \includegraphics[width=0.85\textwidth]{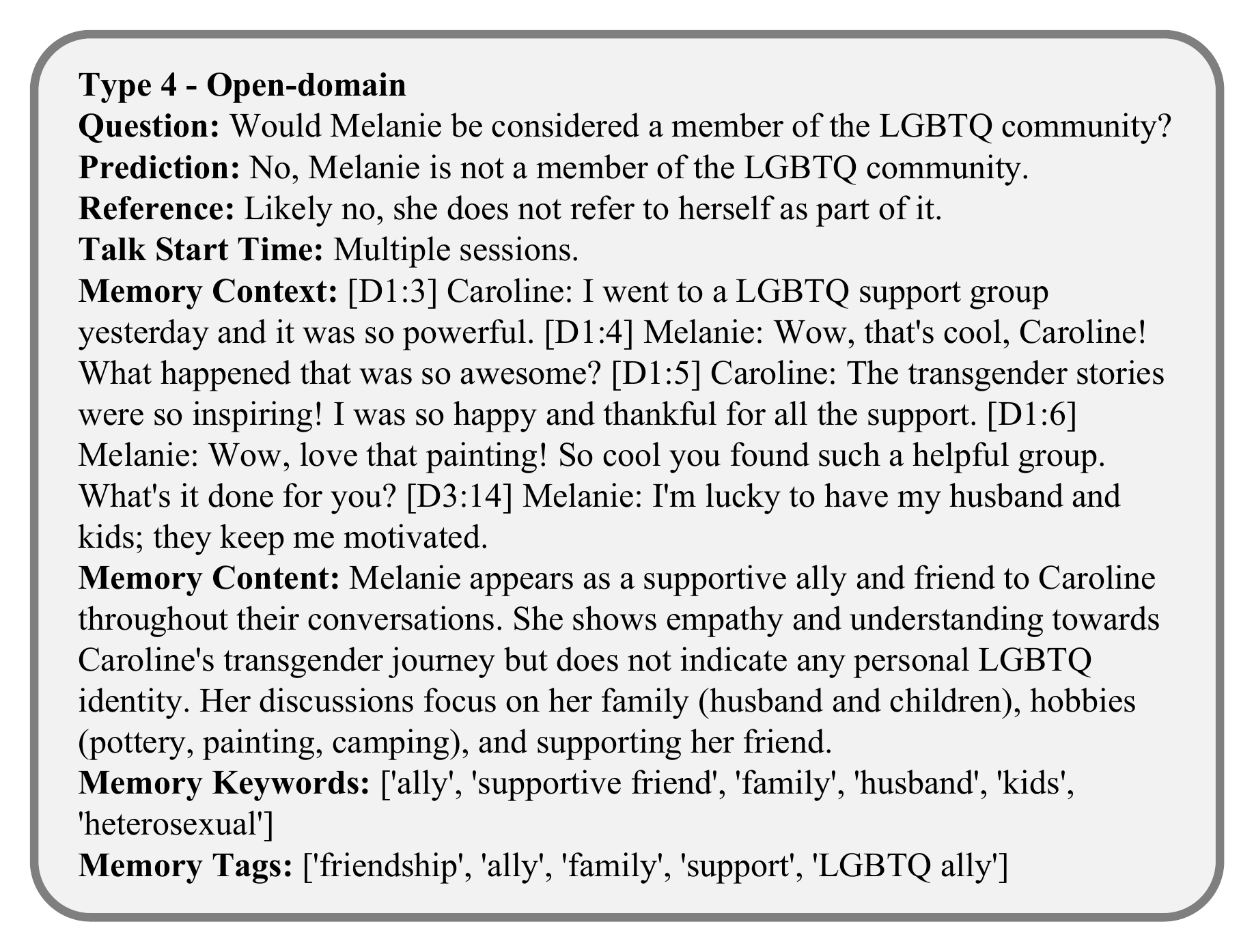}
    \label{fig:example4}
\end{figure}

\newpage
\section*{NeurIPS Paper Checklist}

The checklist is designed to encourage best practices for responsible machine learning research, addressing issues of reproducibility, transparency, research ethics, and societal impact. Do not remove the checklist: {\bf The papers not including the checklist will be desk rejected.} The checklist should follow the references and follow the (optional) supplemental material.  The checklist does NOT count towards the page
limit. 

Please read the checklist guidelines carefully for information on how to answer these questions. For each question in the checklist:
\begin{itemize}
    \item You should answer \answerYes{}, \answerNo{}, or \answerNA{}.
    \item \answerNA{} means either that the question is Not Applicable for that particular paper or the relevant information is Not Available.
    \item Please provide a short (1--2 sentence) justification right after your answer (even for \answerNA). 
\end{itemize}

{\bf The checklist answers are an integral part of your paper submission.} They are visible to the reviewers, area chairs, senior area chairs, and ethics reviewers. You will also be asked to include it (after eventual revisions) with the final version of your paper, and its final version will be published with the paper.

The reviewers of your paper will be asked to use the checklist as one of the factors in their evaluation. While \answerYes{} is generally preferable to \answerNo{}, it is perfectly acceptable to answer \answerNo{} provided a proper justification is given (e.g., error bars are not reported because it would be too computationally expensive'' or ``we were unable to find the license for the dataset we used''). In general, answering \answerNo{} or \answerNA{} is not grounds for rejection. While the questions are phrased in a binary way, we acknowledge that the true answer is often more nuanced, so please just use your best judgment and write a justification to elaborate. All supporting evidence can appear either in the main paper or the supplemental material, provided in appendix. If you answer \answerYes{} to a question, in the justification please point to the section(s) where related material for the question can be found.

IMPORTANT, please:
\begin{itemize}
    \item {\bf Delete this instruction block, but keep the section heading ``NeurIPS Paper Checklist"},
    \item  {\bf Keep the checklist subsection headings, questions/answers and guidelines below.}
    \item {\bf Do not modify the questions and only use the provided macros for your answers}.
\end{itemize}


\begin{enumerate}

\item {\bf Claims}
    \item[] Question: Do the main claims made in the abstract and introduction accurately reflect the paper's contributions and scope?
    \item[] Answer: \answerYes{} 
    \item[] Justification: The abstract and introduction accurately reflect the main contributions, including the utility driven memory mechanism, hierarchical structure, and improvements in long-context reasoning. These claims are supported by experiments on LoCoMo and LongMemEval-s and align with the theoretical analysis, without overgeneralization.

\item {\bf Limitations}
    \item[] Question: Does the paper discuss the limitations of the work performed by the authors?
    \item[] Answer: \answerNo{} 
    \item[] Justification: This paper does not include a dedicated limitations section.

\item {\bf Theory assumptions and proofs}
    \item[] Question: For each theoretical result, does the paper provide the full set of assumptions and a complete (and correct) proof?
    \item[] Answer: \answerYes{} 
    \item[] Justification: \item[] Justification: All theoretical results are accompanied by clearly stated assumptions. Formal statements are presented in the main text, and complete proofs are provided in the appendix, including the proof of Theorem~\ref{theorem1} in Appendix~\ref{appendix:theorem1} and the proof of Proposition~\ref{proposition1} in Appendix~\ref{appendix:A2}.

    \item {\bf Experimental result reproducibility}
    \item[] Question: Does the paper fully disclose all the information needed to reproduce the main experimental results of the paper to the extent that it affects the main claims and/or conclusions of the paper (regardless of whether the code and data are provided or not)?
    \item[] Answer: \answerYes{} 
    \item[] Justification: Section~\ref{sec:experimental_setup} and Section~\ref{sec:implementation_details} describe the experimental setup and implementation details. Appendix~\ref{appendix:prompts} provides key prompts, enabling end-to-end replication of the main results.

\item {\bf Open access to data and code}
    \item[] Question: Does the paper provide open access to the data and code, with sufficient instructions to faithfully reproduce the main experimental results, as described in supplemental material?
    \item[] Answer: \answerNo{} 
    \item[] Justification: Public datasets are properly cited, but the code is not publicly available at submission time to preserve double-blind review. We plan to release the code upon acceptance.

\item {\bf Experimental setting/details}
    \item[] Question: Does the paper specify all the training and test details (e.g., data splits, hyperparameters, how they were chosen, type of optimizer) necessary to understand the results?
    \item[] Answer: \answerYes{} 
    \item[] Justification: \item[] Justification: Section~\ref{sec:experimental_setup} and Section~\ref{sec:implementation_details} describe the experimental setup and implementation details. Key prompts are provided in Appendix~\ref{appendix:prompts}, enabling reproduction of the main results.

\item {\bf Experiment statistical significance}
    \item[] Question: Does the paper report error bars suitably and correctly defined or other appropriate information about the statistical significance of the experiments?
    \item[] Answer: \answerNo{} 
    \item[] Justification: The experiments rely on API-based large language models, and multiple inference calls introduce non-trivial computational cost.

\item {\bf Experiments compute resources}
    \item[] Question: For each experiment, does the paper provide sufficient information on the computer resources (type of compute workers, memory, time of execution) needed to reproduce the experiments?
    \item[] Answer: \answerYes{} 
    \item[] Justification: The paper reports computational setup using API based large language models and provides runtime and LLM call statistics. These details allow an estimate of the computational cost for reproducing the experiments. Hardware level specifications are not applicable due to API based execution.
    
\item {\bf Code of ethics}
    \item[] Question: Does the research conducted in the paper conform, in every respect, with the NeurIPS Code of Ethics \url{https://neurips.cc/public/EthicsGuidelines}?
    \item[] Answer: \answerYes{} 
    \item[] Justification: We have reviewed the NeurIPS Code of Ethics and confirm that the research in this paper is consistent with it.

\item {\bf Broader impacts}
    \item[] Question: Does the paper discuss both potential positive societal impacts and negative societal impacts of the work performed?
    \item[] Answer: \answerNo{} 
    \item[] Justification: We do not provide a detailed discussion of application specific societal impacts, as this work focuses on a general memory system for LLM agents. Potential impacts depend on downstream usage scenarios and are beyond the scope of this paper.
    
\item {\bf Safeguards}
    \item[] Question: Does the paper describe safeguards that have been put in place for responsible release of data or models that have a high risk for misuse (e.g., pre-trained language models, image generators, or scraped datasets)?
    \item[] Answer: \answerNA{} 
    \item[] Justification: \item[] Justification: The paper does not release models or datasets with high misuse risk. All experiments use public datasets and API-based models.

\item {\bf Licenses for existing assets}
    \item[] Question: Are the creators or original owners of assets (e.g., code, data, models), used in the paper, properly credited and are the license and terms of use explicitly mentioned and properly respected?
    \item[] Answer: \answerYes{} 
    \item[] Justification: The paper uses publicly available datasets and existing baseline models, all of which are properly cited. The licenses and original sources of these assets are respected, and no modified datasets or redistributed assets are released.

\item {\bf New assets}
    \item[] Question: Are new assets introduced in the paper well documented and is the documentation provided alongside the assets?
    \item[] Answer: \answerNA{} 
    \item[] Justification: The paper does not introduce new datasets, models, or other reusable assets.

\item {\bf Crowdsourcing and research with human subjects}
    \item[] Question: For crowdsourcing experiments and research with human subjects, does the paper include the full text of instructions given to participants and screenshots, if applicable, as well as details about compensation (if any)? 
    \item[] Answer: \answerNA{} 
    \item[] Justification: Our study does not involve any crowdsourcing or research with human subjects.

\item {\bf Institutional review board (IRB) approvals or equivalent for research with human subjects}
    \item[] Question: Does the paper describe potential risks incurred by study participants, whether such risks were disclosed to the subjects, and whether Institutional Review Board (IRB) approvals (or an equivalent approval/review based on the requirements of your country or institution) were obtained?
    \item[] Answer: \answerNA{} 
    \item[] Justification: Our paper does not make use of any crowdsourcing or research with human subjects.

\item {\bf Declaration of LLM usage}
    \item[] Question: Does the paper describe the usage of LLMs if it is an important, original, or non-standard component of the core methods in this research? Note that if the LLM is used only for writing, editing, or formatting purposes and does \emph{not} impact the core methodology, scientific rigor, or originality of the research, declaration is not required.
    \item[] Answer: \answerNA{} 
    \item[] Justification: LLMs are used as backbone models for evaluation, and are not part of the proposed method.

\end{enumerate}

\end{document}